\documentclass{article}
\usepackage{moreverb,url,flushend}
\usepackage{algorithm,cite}
\usepackage[noend]{algpseudocode}

\makeatletter
\def\BState{\State\hskip-\ALG@thistlm}
\makeatother

\usepackage{MnSymbol,multirow}
\usepackage{amsmath,amsfonts,tikz,amsthm}
\newtheorem{theorem}{Theorem}[section]
\newtheorem{corollary}{Corollary}
\newtheorem{lemma}{Lemma}
\newtheorem{proposition}{Proposition}
\newtheorem{example}{Example}
\newtheorem{remark}{Remark}
\newtheorem{definition}{Definition}

\newcommand\A{\mathrm{AlgorithmAllocation}}
\newcommand\SMU{\mathrm{SumMemoryUsage}}
\newcommand\MU{\mathrm{MemoryUsage}}
\newcommand\TO{\mathrm{TotalOutput}}

\usepackage{jfrc}
\usepackage{graphicx}
\usepackage{apalike}
\usepackage{setspace}
\usepackage[colorlinks,bookmarksopen,bookmarksnumbered,citecolor=red,urlcolor=red]{hyperref}
\usepackage{subfiles}

\providecommand{\keywords}[1]
{
  \small	
  \textbf{\textit{Keywords---}} #1
}

\title{Optimal Algorithm Allocation for Robotic Network Cloud Systems}

\author{
Saeid Alirezazadeh\thanks{C4 - Estrada Municipal, 506, 6200-284 Covilh\~{a}, PORTUGAL} \\
C4-Cloud Computing Competence Center\\ 
Universidade da Beira Interior\\
Covilh\~{a}, Portugal\\
\texttt{saeid.alirezazadeh@gmail.com}\\
\And
Andr\'{e} Correia\\
C4-Cloud Computing Competence Center\\ 
Universidade da Beira Interior\\
Covilh\~{a}, Portugal\\
\texttt{andrerspcorreia@gmail.com} \\
\And
Lu\'{i}s A. Alexandre \\
NOVA LINCS\\
Universidade da Beira Interior\\
Covilh\~{a}, Portugal\\
\texttt{lfbaa@di.ubi.pt} \\
}

\begin{document}
\maketitle

\begin{abstract}
A robotic network is a system with multiple robots connected by a communication network. Certain tasks that cannot be accomplished with available robotic resources are candidates for the use of cloud robotics, which overcomes the limitations of the robot network by adding to the network, either local or remote servers or cloud infrastructure, to aid in computational demanding tasks or storage. Previous studies have mainly focused on minimizing the cost of the robots in retrieving resources by knowing the resource allocation in advance. We develop a method for a robotic network cloud system that includes robots, fog and cloud nodes, to determine where each algorithm should be allocated so that the system achieves optimal performance, regardless of which robot initiates the request. We can find the minimum required memory for the robots and the optimal way to allocate the algorithms with the shortest time to complete each task. We experimentally compare our method with a state-of-the-art method, using real-world data, showing the improvements that can be obtained.
\end{abstract}


\section{Introduction}

The capabilities of robots and the number of tasks they can perform are limited, which means that some tasks exceed the capabilities of a single robot. In these cases, a natural solution is to use multiple robots that cooperate with each other to perform the task, rather than having a single robot perform a task. Usually these cooperating robots are connected through a network, hence the system is called a robotic network. Robotic networks have been widely studied, and their main applications are described in \cite{osumi:2014,michael:2012,Hu:2012,McKee:2008,Mckee:2000,Tenorth:2013,Kamei:2012}, among others. 

As mentioned in \cite{Hu:2012}, the capacity of a robotic network is higher than that of a single robot, but it is still limited by the capacities of all robots. Therefore, some tasks may exceed the capacity of a robotic network, i.e., tasks related to human-robot interaction, such as speech (\cite{jelinek:1997}), face (\cite{jain:2011}), and object (\cite{xu:2013}) recognition, which are very computationally intensive. To solve this problem, we can increase the number of robots to increase the capacity, but in this case we also increase the complexity of the model.

Cloud robotics is described as a way to circumvent some of the computational limitations of robots by using the Internet and cloud infrastructure to delegate computation and share big data in real-time \cite{kehoe:2015}. An important factor in determining the performance of cloud-based robotic systems is deciding whether a task should be uploaded to the cloud, processed on a server (fog computing \cite{bonomi:2012}) or executed on one of the robots (edge computing \cite{shi:2016}), the so-called, allocation problem. 

The most recent review of works on cloud robotics, \cite{saha:2018}, which includes works from 2012 to 2018, found that task allocation is still a challenge. And the works that address the task allocation problem, such as \cite{wang:2014,wang:2016}, and \cite{kong:2017}, are mainly concerned with optimizing the cost for the robots to fetch all the necessary resources, assuming that the resources (algorithms and data) are allocated in advance. But these studies do not answer the question of how to proceed with optimal resource allocation. However, if the resource allocation is determined in advance, they describe methods to reduce costs.

In a multi-robot system, let $T$ be a finite set of tasks that the system can perform. As tasks arrive sequentially, a time segment can be defined as the time interval between two consecutive new arriving tasks. In a time segment, the system executes a set of tasks, $T_1$, which is a subset of the set of all tasks. At the same time, a new set of tasks, $T_2$, arrives to be executed. As shown in Figure~\ref{fig0}, there are two types of task allocation:
\begin{itemize} 
\item \textbf{static task allocation}, which provides a solution to the question of achieving the optimal performance of the system by assigning the tasks in the set of all tasks, $T$.
\item and \textbf{dynamic task allocation}, which answers the question of achieving the optimal performance by considering the order of the set of arrived tasks with respect to the arrival time, $(T_i)_{i\in\mathbb{N}}$, and assigning them dynamically.
\end{itemize}
Each task can be splitted into multiple algorithms, which must be executed, simetimes simultaneously, to be able to complete the task. Each algorithm, that may require certain inputs, will process those inputs, and provide certain outputs. The information required to perform a task is called the inputs to the task. For example, to grasp an object with an arm, the pose of the arm and of the object are inputs for grasping, which are collected by applying some algorithms to the sensory information.

Static task allocation can be considered the primary evaluation of a cloud robotic system. In dynamic task allocation, a set of tasks arrives in the system and we need to decide which task should be assigned to which node of the cloud robotic system. While a task is assigned to a node, that node may send requests to execute the required algorithms to other nodes to complete the task. In order to reduce the memory usage and time required to complete the task, it is sometimes better to execute these algorithms on nodes other than the one to which the task is assigned. Static allocation is about finding the optimal way to allocate all the algorithms needed to execute all the tasks, in the sense of minimizing the execution cost of the tasks. Static allocation solves the question of how to optimally distribute all the algorithms among the nodes. It does this by ensuring that all required information is optimally collected by the node to which the task is assigned. Thus, static task allocation yields the minimum cost of the system after all tasks have been executed, while dynamic task allocation yields the minimum cost of the system in a given time horizon. Moreover, the minimum cost obtained by static task allocation is the largest lower bound on the minimum cost obtained by dynamic task allocation for the same set of tasks. This is because in static task allocation we minimize the cost of performing each task, regardless of where it is assigned. Using this result in dynamic task allocation means that the execution cost for each task is already minimized and we only need to minimize other costs. In other words, static task allocation is not only as important as dynamic task allocation, but it also shows how to achieve the optimal performance of cloud robotic systems in performing each task. In this paper, we focus on the static allocation of algorithms that simultaneously minimize the memory usage of robots and the total execution time of all tasks.

This paper is organized as follows: First, we briefly describe previous studies dealing with the allocation problem. In the following section, we describe the time optimization model for static algorithm allocation for a single-robot cloud system. We then generalize it to a multi-robot cloud system. Next, we develop a theory for minimizing the maximum memory usage of all robots. We propose a method that simultaneously minimizes the total time to obtain all outputs of all algorithms from all robots and the maximum memory usage of robots. Next, we present some simple examples to explain how each part of the proposed method works, and finally, we provide a real-world experiment to test how the result of our proposed method compares to the result of the state-of-the-art algorithm.

\section{Related Works}

In static task allocation, \cite{Lin:1995} studied static algorithm allocation for a multi-robot system without considering cloud infrastructure and communication times. As far as we know, the only works that have addressed a similar problem are \cite{li:2018}, and \cite{Ours:2020}. Alirezazadeh and Alexandre in \cite{Ours:2020} solved the allocation problem by minimizing memory and time for a single robotic cloud system. They described the graph of all algorithms as a semi-lattice with additional virtual algorithms that determine the priority order of the data. Then they described an algebra of memory and time, which provides the effect of parallel and serial algorithms on execution time and memory usage. Since the problem addressed in \cite{Ours:2020} is for a single robot, the optimization model is not formulated and an immediate solution is provided using a combinatorial method. Li et~al.~in \cite{li:2018} approached the allocation problem by considering only the minimum time and ignoring the memory parameter, and by minimizing the total execution time without fully considering communications. The result in \cite{li:2018}, as shown in \cite{Ours:2020}, provides an optimal solution for the cases where the average execution times of all algorithms are much larger than the average communication times between nodes, or for the cases where the average communication times between nodes are negligible.
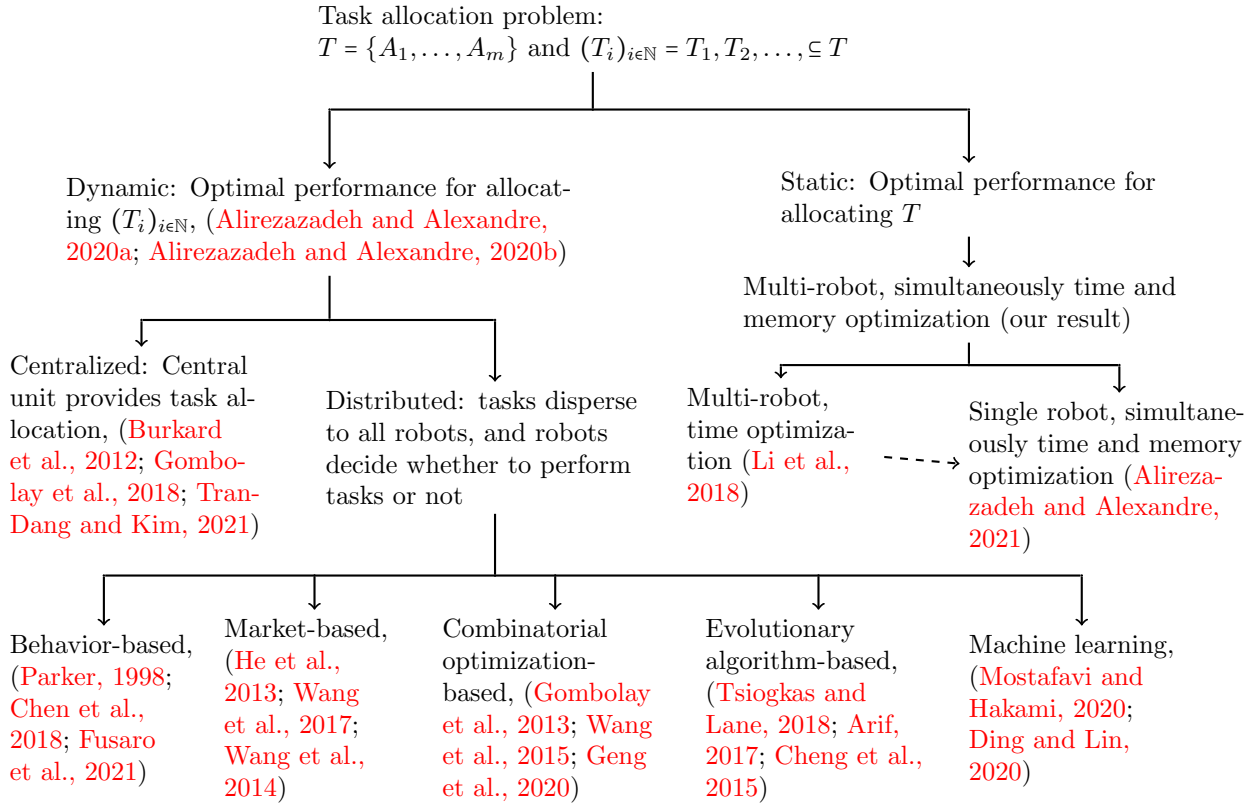
\begin{figure*}[tb]\centering
\begin{tikzpicture}
\begin{scope}%[every node/.style={rectangle,thick,draw,rounded corners=.8ex}]
    \node (A) at (0,0) [text width=7.3cm] {Task allocation problem: \\$T=\{A_1,\ldots,A_m\}$ and  $(T_i)_{i\in\mathbb{N}}=T_1,T_2,\ldots, \subseteq T$};
    \node[fill,circle,scale=0.1]  (B) at (0,-1) {};
    \node[fill,circle,scale=0.1]  (C) at (-3.5,-1) {};
    \node[fill,circle,scale=0.1]  (D) at (5,-1) {};
    \node (E) at (-3.5,-2.5)  [text width=7cm] {Dynamic: Optimal performance for allocating $(T_i)_{i\in\mathbb{N}}$, \cite{Ours:2020h,Ours:2020imp}};
    \node (F) at (5,-2.2)  [text width=5cm] {Static: Optimal performance for allocating $T$};
    \node[fill,circle,scale=0.1]  (G) at (-3.5,-3.8) {};
    \node[fill,circle,scale=0.1]  (H) at (-6,-3.8) {};
    \node[fill,circle,scale=0.1]  (I) at (-1.3,-3.8) {};
    \node (J) at (-6,-5.5) [text width=3.5cm]{Centralized: Central unit provides task allocation, \cite{Burkard:2012,Gombolay:2018,Dang:2021}};
    \node (K) at (-1.3,-5.5)[text width=4.5cm] {Distributed: tasks disperse to all robots, and robots decide whether to perform tasks or not};
    \node[fill,circle,scale=0.1]  (L) at (-1.3,-7.2) {};
    \node[fill,circle,scale=0.1]  (M) at (-3.7,-7.2) {};
    \node[fill,circle,scale=0.1]  (N) at (-6.5,-7.2) {};
    \node[fill,circle,scale=0.1]  (O) at (-0.5,-7.2) {};
    \node[fill,circle,scale=0.1]  (P) at (3,-7.2) {};
    \node[fill,circle,scale=0.1]  (a) at (6.5,-7.2) {};
    \node (Q) at (-6.5,-9) [text width=2.5cm]{Behavior-based, \cite{Parker:1998,Chen:2018,Fusaro:2021}};
    \node (R) at (-3.7,-9)[text width=2.4cm] {Market-based, \cite{Alaa:2013,Wang:2017,wang:2014}};
    \node (S) at (-0.5,-9)[text width=3cm] {Combinatorial optimization-based, \cite{Gombolay:2013,Wang:2015,Geng:2020}};
    \node (T) at (3,-9) [text width=3cm]{Evolutionary algorithm-based, \cite{Lane:2018,Arif:2017,Cheng:2015}};
    \node  (U) at (5,-3.6) [text width=6cm]{Multi-robot, simultaneously time and memory optimization (our result)};
    \node (b) at (6.5,-9) [text width=3cm]{Machine learning, \cite{mostafavi:2020,Ding:2020}};
    \node[fill,circle,scale=0.1]  (V) at (5,-4.4) {};
    \node[fill,circle,scale=0.1]  (W) at (7,-4.4) {};
    \node[fill,circle,scale=0.1] (X) at (2.5,-4.4) {};
    \node (Y) at (2.5,-5.5) [text width=2.5cm] {Multi-robot, time optimization \cite{li:2018}};
    \node (Z) at (7,-5.9) [text width=4cm] {Single robot, simultaneously time and memory optimization \cite{Ours:2020}};

\end{scope}

    \path [->] (A) edge[thick,-] (B);
    \path [->] (C) edge[thick,-] (D);
    \path [->](C) edge [thick,->] (E);
    \path [->](D) edge [thick,->] (F);
    \path [->](E) edge [thick,-] (G);
    \path [->](F) edge [thick,->] (U);
    \path [->](U) edge [thick,-] (V);
    \path [->](I) edge [thick,-] (H);
    \path [->](W) edge [thick,-] (V);
    \path [->](H) edge [thick,->] (J);
    \path [->](I) edge [thick,->] (K);
    \path [->](W) edge [thick,-] (X);
    \path [->](K) edge [thick,-] (L);
    \path [->](N) edge [thick,-] (a);
    \path [->](a) edge [thick,->] (b);

    \path [->](N) edge [thick,->] (Q);
    \path [->](M) edge [thick,->] (R);
    \path [->](O) edge [thick,->] (S);
    \path [->](P) edge [thick,->] (T);
    \path [->](X) edge [thick,->] (Y);
    \path [->](W) edge [thick,->] (Z);
    \path [->](Y) edge [thick, dashed,->] (Z);
\end{tikzpicture}
\caption{Overview of the task allocation methods. The dashed arrow is used only to show that the result in \cite{Ours:2020} for a single robot is not a specific instance of the result of \cite{li:2018}. From our result, we can extract the result of \cite{li:2018}.}
\label{fig0}
\end{figure*}

As shown in Figure~\ref{fig0}, dynamic task allocation is divided into centralized and distributed allocation. In centralized methods, a central planning unit that has information about the entire environment handles task allocation, \cite{Burkard:2012}. In distributed allocation, instead of a central unit, all tasks are distributed to all robots, and robots decide which tasks to perform without central coordination. Five main approaches have been used for distributed allocation:
\begin{itemize}
\item behavior-based: uses problem features to decide whether or not a robot should consider a task, \cite{Parker:1998};
\item market-based: relies on an auction-based mechanism, \cite{Alaa:2013};
\item combinatorial optimization-based: transforms the problems into combinatorial optimization problems and uses an appropriate existing technique to solve them, \cite{Gombolay:2013};
\item evolutionary algorithm-based: uses evolutionary operators on the space of solutions, \cite{Lane:2018,Arif:2017}.
\item machine learning algorithm: uses a machine learning algorithm to find an optimal task allocation, \cite{mostafavi:2020,Ding:2020}.
\end{itemize}
For task allocation in a cloud robotic system, most studies only move computationally intensive tasks to the cloud without considering the communication time between robots and the cloud infrastructure, \cite{davinci:2010,Slam:2015,Rapyuta:2013}.

Dynamic task allocation has been widely studied. The most recent works are: \cite{Cheng:2015} that proposed an architecture for wearable computing devices and study offloading strategies that preserve the required delays due to the characteristics of the devices, and proposed a fast algorithm based on the genetic algorithm as an offloading strategy; \cite{Chen:2018} that proposed an algorithm to optimize the latency, energy consumption, and computational cost considering the characteristics of cloud robotic architecture, task properties, and quality of service problems for cloud robotics; \cite{Gombolay:2018} presented a centralized algorithm, called ''Tercio'', for task allocation by minimizing latency and physical proximity to tasks; \cite{Wang:2015} investigated the resource sharing problem in cloud robotic systems and presented a resource management strategy with near real-time data retrieval; \cite{Wang:2017} studied real-time resource allocation based on the hierarchical auction-based method; \cite{Schillinger:2018} proposed a method for centralized dynamic task allocation based on modeling the problem as a non-deterministic finite automaton, and a minimum cost solution is presented depending on the energy level of the robots; \cite{Zhang:2018} proposed a method to handle the deadline of a task and also minimize the total cost; \cite{Chen:2019} studied task allocation under the assumption that the number of robots changes and the tasks are changeable; \cite{WANG:2020} studied the optimal task allocation for the case where two collaborative robot teams jointly perform some complex tasks, and provided the formulation of the optimization statement of their model based on set-theoretic parameters; \cite{Ours:2020imp} proposed a load balancing algorithm for cloud robotic systems with time window constraints, and used the time windows to generate the grid of tasks and the task allocation is then done by the load balancing algorithm; and \cite{Ours:2020h} translated the optimal task allocation into finding the maximum volume of a subspace of a hyperspace. They investigated the compatibility of a node to perform a task, communication and communication instability, and the capability of fog, cloud, and robots; \cite{mostafavi:2020} proposed a scheduling method to minimize the makespan and response time and increase resource efficiency in a cloud infrastructure. The scheduling method is based on reinforcement learning. At each time step, the size of the occupied buffer and total task length of virtual machines are considered independent to use Bayes' theorem, and the $Q$-values are estimated; \cite{Geng:2020} developed a method for simultaneously minimizing resource usage, time, and cost, and performing load balancing in a cloud infrastructure. The method consists of formulating all the objectives, and using the hybrid angle strategy \cite{Tseng:2006} to find the optimal solution; \cite{emam:2020} studied minimizing the cost function in a dynamic environment where task priorities may change due to changes in the environment. The optimal allocation is achieved by updating the states of the robots over time and updating the priorities with the changes in the environment; \cite{Brown:2020} studied the allocation and routing of robots to move objects between stations in such a way that the makespan (the time from start to finish) is minimal. The problem is translated into a problem called precedence constrained multi-agent task allocation and path-finding (PC-TAPF), which is an optimization problem to minimize the makespan, and hierarchical algorithm is proposed to find the optimal makespan; \cite{Dang:2021} proposed a scheduling method in which the fog layer handles all the tasks by distributing and balancing the loads among all the fog nodes and reducing the delay. The authors translate the problem into an optimization problem that minimizes the service delivery delay (the time interval between when the fog receives a request and when the IoT node sending the request receives a response); \cite{Fu:2021} proposed a method to decide whether to assign subtasks to edge devices or offload them to fog nodes, so that the total execution time of all subtasks and the total energy consumption of all edge devices are minimal. In this method, the tasks that need to be executed by edge devices are translated into a DAG,  generated by the dependencies of the subtasks. Considering the average execution time and energy consumption of all subtasks on each core of the system architecture, the problem is then transformed into an optimization problem whose solution allows optimal scheduling of the subtasks; \cite{Malencia:2021} has studied a fair redundant allocation of agents to tasks that improves task performance. The proposed method consists of translating tasks and agents into a bipartite graph whose edges are weighted according to the cost of each task assigned to an agent. The redundant allocation problem is about which task should be assigned an additional resource. The problem is transformed into an optimization problem, and an attempt is made to solve it; \cite{Sahni:2021} proposed a mathematical model for joint offloading of multiple tasks that takes into account the dependencies between the subtasks and schedules the network flows in such a way that the completion time of the tasks is minimized. The network flow dependencies are the problem in offloading tasks to multiple devices, leading to competition for bandwidth usage. In the proposed model, the problem is translated into an optimization problem and a solution called Joint Dependent Task Offloading and Flow Scheduling Heuristic (JDOFH) is proposed; \cite{Ding:2020} has proposed a method to find an optimal solution for distributing tasks to servers that minimizes the cost to the user. In the proposed method, the task allocation problem is translated into a reinforcement learning problem where the reward function is the negative average user cost; \cite{Fusaro:2021} proposed a method to minimize the cost of allocating tasks to the human-robot team by translating the problem into a behavioral tree by dividing the tasks into a series of parallel tasks. Given a set of tasks that an agent can perform, the Mixed-Integer Linear Programming (MILP) problems are solved recursively to minimize the cost.

Our method for finding an optimal allocation of algorithms yields two classes for each memory and time up to certain values, the intersection of which yields the optimal solution for minimizing the robot's Memory-Time problem. We begin with a brief description of the objectives, formulate the problem, and then present the solution. Then we illustrate the model by finding solutions considering simulated data. Finally, we use real-world data and provide solutions to the allocation problem.

For solving the allocation problem in terms of the time parameter, we complete the method in \cite{li:2018} by fully considering the communication time. For solving the allocation problem in terms of the memory parameter, we use the notation in \cite{Ours:2020} and then develop a theory for solving the allocation problem by optimizing the memory usage of all robots. To begin, we assume that the robots are identical and none has an advantage over the other. The main advantage of this condition is that if one robot malfunctions, we can easily assign its task to other functioning robots.

Our main contributions are as follows:
\begin{itemize}
\item We complete the optimization problem with respect to the time parameter described in \cite{li:2018} by fully considering the communication time.
\item We developed a theory for task allocation that minimizes only the memory usage of all robots.
\item We extend the optimization problem to minimize both the task completion time and the memory usage of the robots.
\item We analyze the scalability of the proposed method.
\item We test and compare the results of the proposed method with a state-of-the-art method using real-world data and simulations.
\end{itemize}

Throughout this paper, the communication times between the edge, the fog, and the cloud are considered as average communication times since random delays are present. The execution time of an algorithm on the edge, the fog and the cloud is considered as the average execution time of the algorithm on the edge, the fog, and the cloud.

We start by constructing a general theoretical model. Based on that, we formulate the problem and find a set of possible solutions.
\section{Proposed Method}
\subsection{Time Optimization}
We start by considering a single robot cloud system and then extend it to robotic networks.

We first recall the semi-lattice model for graph of all algorithms described in \cite{Ours:2020}.Consider a simplified graph of a cloud robotics architecture, a graph $V_n$ with three nodes, where each node represents either the set of all edges nodes, fog nodes, or the cloud nodes with communication links between them. Consider also the graph of algorithms, that is a graph $V_t$ with algorithm execution dependency drawn with downward edges. The virtual vertices $\mathbf{0}$ and $\mathbf{1}$ are also added to the graph ($\mathbf{1}$ is added at the top of the graph, as a virtual source for all algorithms, and $\mathbf{0}$ is added at the bottom of the graph, as the virtual sink for all algortihms ) to generate a semi-lattice. For more details about the graph of algorithms see \cite{Ours:2020}.

Two nodes in a cloud robotics architecture are called neighbors if they are within a communication range. Equivalently, two nodes are neighbors if they can communicate directly with each other. For more details about the graph of a cloud robotics architecture, see \cite{Hu:2012}. 

Of all possible allocations, we want to minimize the maximum time that an edge node takes between sending requests for the execution of any algorithm and receiving the response. Note that the maximum time is always for the last algorithm (virtual algorithm $\mathbf{0}$ at the bottom of the graph of algorithms).

We refer to the response time of an algorithm initiated at a node, say $\alpha$, as the sum of the following times: 
\begin{itemize} 
\item the average time required to send a request from node $\alpha$ to node $\beta$, where the algorithm is located;
\item the average time taken by the node $\beta$ to gather all the necessary information to execute the algorithm; 
\item the average time it takes for the algorithm to execute on node $\beta$; 
\item the average time it takes for the result of the algorithm to be sent from node $\beta$ to node $\alpha$. 
\end{itemize}
Note that we consider the average time because the execution time and the data transmission time are not exact and may contain some random delays.

The average time taken by the node $\beta$ to gather all the necessary information to execute the algorithm is the sum of the following times:
\begin{itemize}
\item the maximum response time of all predecessors of the algorithm at node $\beta$ in the graph of algorithms;
\item the time taken to receive all additional data from their respective nodes, if there are any. 
\end{itemize}
This model is recursive and assumes initiation by an edge node for any arbitrary allocation of algorithms. Since the graph of algorithms is bounded, and the number of nodes of the cloud system is also bounded, finding the optimal allocation is solvable.

We follow a similar notation as in the problem formulation in \cite{li:2018}:
\begin{itemize}
\item $t_i^s$ is the time when the algorithm $i$ started (start time),
\item $t_i^f$ is the time when the execution of algorithm $i$ finished (response time),
\item $V_n$ is the set of nodes of the cloud robotics architecture including all nodes in the edge, nodes in the cloud (and nodes in the fog, but this is not assumed in \cite{li:2018}),
\item $x_{ik}$ indicates whether node $k$ in $V_n$ has an algorithm $i$ associated with it,
\item the nodes $e$, $f$ and $c$ in $V_n$ denote the node from edge node, fog node and cloud node, respectively;
\item $V_t$ is the set of all algorithms in the graph of dependency of algorithms with additional virtual nodes,
\item $Z_{ik}$ is an indicator that an algorithm $i$ can be assigned to node $k$ in $V_n$, which can accomodate prior information about the places where algorithms need to be executed,
\item $\mathrm{pred}_i$ is the set of algorithms that must be executed before algorithm $i$ is executed (the set of all algorithms in the graph for which algorithm $i$ is their successor),
\item $(k,l)\in E_n$ indicates that nodes $k$ and $l$ are neighbors in the cloud robotics architecture, and $E_n$ is the set of all neighbors,
\item $S_{ji}$ is the size of the intermediate data obtained from algorithm $j$ that must be transmitted to the node executing algorithm $i$ to be used as input to $i$,
\item$y_{ij}=\left\{\begin{array}{ll}1&,~~\text{if task}~ i\in V_t~\text{is scheduled to run after}\\&~~j\in V_t,\\
0&,~~\text{otherwise.}\end{array}\right.$
\item $R_{ik}$ is the runtime of algorithm $i$ on node $k$.
\end{itemize} 
\begin{remark}\label{remm1}
Since the time to get the outputs of the final algorithm is the longest compared to the other algorithms, and the time varies depending on where the algorithms are allocated, we look for an allocation of the algorithms that minimizes this time. A request for the final algorithm can be sent from the edge node at any time, say $t_0$, and the time at which the edge node receives the final algorithm's outputs is $t_m^f$. This means that minimization must be performed for duration $t_m^f-t_0$. If $t_0$ changes to $t_0\pm t$ for some $t\in\mathbb{R}$, then $t_m^f$ also changes to $t_m^f\pm t$. So we can consider $t_0=0$.
\end{remark}
According to Remark~\ref{remm1}, the problem is to minimize the time between the initial request by the edge node $e$, until the last algorithm, indexed by $m$, which is the algorithm $\mathbf{0}$, returns its output to the edge node $e$, $\min:~~ t_m^f(=t_m^f(e))$, subject to the condition that each algorithm can only be executed by exactly one of the nodes, i.e. $x_{ic}+x_{if}+x_{ie}=1$, for all $i\in V_t$. The prior knowledge of where the algorithms are executed can be specified as follows:
$$\begin{array}{ll}
x_{ik}\leq Z_{ik},&\forall i\in V_t,~k\in\{e,f,c\}\\
Z_{ic}+Z_{if}+Z_{ie}\leq3,&\forall i\in V_t\\
Z_{ic}+Z_{if}+Z_{ie}\geq1,&\forall i\in V_t\\
Z_{ik}\in\{0,1\},&\forall i\in V_t,~k\in\{e,f,c\}
\end{array}$$
where $Z_{ik}$, which can be $0$ or $1$, is an indicator of whether an algorithm $i$ can be assigned to a set of nodes, $k$ being one of them, and cannot be assigned to the remaining nodes.

The time at which algorithm $i$ is started, is the sum of the following times:
\begin{itemize}
\item the time in which the set of all immediate predecessors of algorithm $i$ are executed:
$$T_{1,i}=\max\limits_{j\in\mathrm{pred}_i}\left\{t_j^f\left(\sum_{p\in \{c,f,e\}}x_{jp}p\right)\right\};$$
\item the time taken to send intermediate data, used as input by algorithm $i$, from the nodes generating that input to the node executing $i$:
$$T_{2,i}^k=\sum\limits_{j\in\mathrm{pred}_i}\mathrm{TransmissionTime}_k(S_{ji}),$$
where $\mathrm{TransmissionTime}_k$ is the average time to transmit $S_{ji}$ data to node $k$. From now on, we denote by $S_{ji}$ any additional information that needs to be transmitted to $i$ as input of algorithm $i$ in addition to the information obtained in the previous step.
\end{itemize}
It follows that, $t_i^s=T_{1,i}+\sum_{k\in\{c,f,e\}}x_{ik}T_{2,i}^k$. Note that in this formulation, we assume that the fog and the cloud each have only one node.

The time at which the algorithm $i$ terminates is the sum of the following times:
\begin{itemize}
\item the time that algorithm $i$ is started, $t_i^s$;
\item the runtime of algorithm $i$ on node $k$;
\item the average time for transmitting the output data of algorithm $i$, whose size is denoted by $\mathrm{OutputSize}_i$, to node $p$ that requested algorithm $i$.
\end{itemize}
Hence, 
$$t_i^f(p)=t_i^s+\sum_{k\in\{c,f,e\}}x_{ik}R_{ik}+\mathrm{TransmissionTime}_p(\mathrm{OutputSize}_i).$$
Note that in order to minimize the overall time of the system we minimize the term $t_m^f(e)$, which means that starting from time $0$, the time in which the execution of algorithm $\mathbf{0}$ is completed is equal to the maximum time required for the robot to obtain all outputs of all algorithms. Also, note that the maximum overall time is always for the last algorithm requested by all robots (virtual algorithm at the bottom of the graph of algorithms).

Altogether, this implies the following formulation for minimizing the time:
\begin{align}\label{optnew}
\min:&~t_m^f\left(=t_m^f(e)\right)\\
\hline
&\nonumber\\
\text{s.t.}:&\sum_{k\in\{e,f,c\}}x_{ik}=1\nonumber\\
&x_{ik}\leq Z_{ik},~~\forall i\in V_t,~k\in\{e,f,c\}\nonumber\\
&1\leq Z_{ic}+Z_{if}+Z_{ie}\leq3,~~\forall i\in V_t\nonumber\\
&t_i^s=T_{1,i}+\sum_{k\in\{c,f,e\}}x_{ik}T_{2,i}^k,~~\forall i\in V_t\nonumber\\
&T_{1,i}=\max\limits_{j\in\mathrm{pred}_i}\left\{t_j^f\left(\sum_{p\in \{c,f,e\}}x_{jp}p\right)\right\},~~\forall i\in V_t\nonumber\\
&T_{2,i}^k=\sum\limits_{j\in\mathrm{pred}_i}\mathrm{TransmissionTime}_k(S_{ji}),~~\forall i\in V_t,~k\in\{e,f,c\}\nonumber\\
&t_i^s\geq\max\limits_{j\in V_t}\left\{\sum\limits_{k\in V_n}x_{ik}x_{jk}y_{ij}\times\left(t_j^f\left(\sum_{p\in \{c,f,e\}}x_{jp}p\right)\right)\right\},~~\forall i\in V_t\nonumber\\
&t_i^f(p)=t_i^s+\sum_{k\in\{c,f,e\}}x_{ik}R_{ik}+\mathrm{TransmissionTime}_p(\mathrm{OutputSize}_i),~~\forall i\in V_t\nonumber\\
&x_{ik},y_{ij},Z_{ik}\in\{0,1\},\nonumber
\end{align}
where $t_j^f(k)$ is the response time of algorithm $j$ when initiated by node $k$, and the other parameters are either the same as in the model of \cite{li:2018} or as described above. The differences are that our model:
\begin{enumerate}
\item sets boundary conditions for $Z_{ik}$ that incorporate the effect of prior information about where we can run algorithms and the cases where we assume, for example, that an algorithm cannot be run on the edge. A non-informative prior can also be described with this model.
\item the initiation and termination of algorithms are now fixed, and all refer to the start of the request from an edge node and the reception of the result in the same node.
\item the transmission of the algorithms' output to the edge nodes is considered, both in the start time and in the response time of the algorithms.
\end{enumerate}

By construction, this model is equivalent to the algebraic model in \cite{Ours:2020} and the only difference is that the model in \cite{Ours:2020} uses algebraic expressions, while here we use analytic expressions. We can extend it to multiple edge nodes. But the formulation needs some modifications.

We assume cooperative robotics. The following problem statement is made by considering additional nodes and a cloud robotics architecture with the robotics network. By considering Remark~\ref{remm1} for all edge nodes, the problem statement~\eqref{optnew} must include data transmission within robots, fog nodes, and cloud nodes, so the problem \eqref{optnew} becomes:
\begin{align}\label{optsev}
\min:&~t_m^f=\sqrt{\sum_{e=1}^E(t_m^f(e))^2}\\
\hline
&\nonumber\\
\text{s.t.}:&\sum_{k\in V_n}x_{ik}=1\nonumber\\
&x_{ik}\leq Z_{ik},~~\forall i\in V_t,~k\in V_n\nonumber\\
&1\leq \sum_{k=1}^{E+C+F}Z_{ik}\leq E+C+F,~~\forall i\in V_t\nonumber\\
&t_i^s=T_{1,i}+\sum_{k\in V_n}x_{ik}T_{2,i}^k,~~\forall i\in V_t\nonumber\\
&T_{1,i}=\max\limits_{j\in\mathrm{pred}_i}\left\{t_j^f\left(\sum_{p\in V_n}x_{jp}p\right)\right\},~~\forall i\in V_t\nonumber\\
&T_{2,i}^k=\sum\limits_{j\in\mathrm{pred}_i}\mathrm{TransmissionTime}_k(S_{ji})~~\forall i\in V_t,~k\in V_n\nonumber\\
&t_i^s\geq\max\limits_{j\in V_t}\left\{\sum\limits_{k\in V_n}x_{ik}x_{jk}y_{ij}\times\left(t_j^f\left(\sum_{p\in V_n}x_{jp}p\right)\right)\right\},~~\forall i\in V_t\nonumber\\
&t_i^f(p)=t_i^s+\sum_{k\in V_n}x_{ik}R_{ik}+\mathrm{TransmissionTime}_p(\mathrm{OutputSize}_i),~~\forall i\in V_t\nonumber\\
&x_{ik},y_{ij},Z_{ik}\in\{0,1\},\nonumber
\end{align}
where $t_j^f(e)$ is the response time of algorithm $j$ when initiated by edge node $e\in E$. The rest of the notations are adapted from \cite{li:2018} and the problem statement~\eqref{optnew}. In this formulation, a request for an algorithm $A$ is sent from a robot. Then necessary algorithms $B_1,\ldots, B_q$ for executing algorithm $A$ are requested from the nodes to which they are assigned. And then
\begin{itemize}
\item if necessary conditions for the execution of algorithm $B_i$ are satisfied, then algorithm $B_i$ is executed, and its results are returned to the node from which it was requested;
\item if necessary conditions for execution of algorithm $B_i$ do not hold, then iterate similarly (requests for necessary algorithms are sent from the node to which algorithm $B_i$ is assigned and on which it cannot be executed to the nodes to which necessary algorithms for the execution of $B_i$ are assigned).
\end{itemize}
\begin{remark}\label{rem:rem0}
In problem statement~\eqref{optsev}, the intuition is that an edge node sends a request to execute an algorithm. After the particular node associated with the algorithm executes the algorithm, the output is returned to the same edge node we started with. We assume that all edge nodes can initiate requests to execute all algorithms. Due to the architecture and neighborhood relationship between the nodes, the final algorithm's response time varies depending on the number of edge nodes and on changing the edge node initiating the request. We need to minimize the total time in which all edge nodes receive the outputs of the final algorithm $m$. To answer the question of how to allocate algorithms to the nodes to optimize the overall performance of the system, we need to consider that the request for an algorithm can be sent from any edge node. So, to minimize the time, we assume that each edge node sends a request for the final algorithm, the virtual algorithm $\mathbf{0}$. For each edge node, the response time of the final algorithm changes depending on the allocation of algorithms in $\mathbb{R}^+$. Now, as we move to all the edge nodes, the respective times change, where the $E$-dimensional space, $\mathbb{R}^E$, obtains the optimal allocation by minimizing the distance of the response times of the final algorithm with respect to the allocation of the algorithms for all the edge nodes, i.e., by minimizing:
$$\sqrt{\sum_{e=1}^E(t_m^f(e))^2}.$$
\end{remark}

One of the most suitable methods to find an optimal solution to the problem~\eqref{optsev} is the branchandbound algorithm, see \cite{mistry:2018}.

\subsection{Memory Optimization}\label{memsec}
We now address the allocation problem concerning the optimal memory usage by edge nodes in a robotic network cloud system with $k$ edge nodes. Suppose that the algorithms are $A=\{A_1,\ldots,A_n\}$ with $\mathcal{ SL }(G(A))$ as the graph of algorithms with downward edges. We follow the notation used in \cite{Ours:2020}.

Recall the algebra of memory described in \cite{Ours:2020}: for an algorithm $A_i$, we describe the memory usage by algorithm $A_i$, with $m_{in}(A_i)$, $m_{pr}(A_i)$, and $m_{ou}(A_i)$ which are the total memory size of the input, processing, and output of algorithm $A_i$, respectively. By $m_{in}(A_i)|_{F\cup C}$, we mean the total size of data necessary to execute algorithm $A_i$ that is not included in the outputs of other algorithms allocated in the edge nodes, and this data is transmitted by some nodes in the fog or cloud, and $m_{in}(A_i)|_E$ is the complement of $m_{in}(A_i)|_{F\cup C}$. Since all edge nodes can send requests to execute any algorithm, $m_{ou}(A_i)$ must be considered in each edge node. Therefore, there must be a fixed amount of memory available in each edge node equal to the total memory required to store all algorithms' outputs, $\TO (A_i)$.

If an algorithm $A_i$ is allocated to an edge node $R_j$, then the edge node $R_j$ must contain the processing and input memory of the algorithm $A_i$, that is, $m_{pr}(A_i)$ and $m_{in}(A_i)|_{F\cup C}$. Thus, for each edge node $R_j$, with $j=1,\ldots,k$:
\begin{align}\label{eq:eq3}
\MU(R_j,\A)=&\TO(A)+TM(R_j)+\sum_{i=1}^nx_{i,j}\left(m_{pr}(A_i)+m_{in}(A_i)\right),
\end{align}
where $TM(R_j)$ is the size of the data transmitted to other nodes (some information such as a graph of the architecture, the node sending a request for an algorithm and the node that needs to execute the algorithm, etc.) and $x_{i,j}$ indicates whether an algorithm $i$ is assigned to the node $R_j$.

Our goal is not only to minimize the total memory usage of the edge nodes with respect to $\A$ ($\A$ is defined as all possible algorithm allocations to all nodes of the robotic network cloud system, which can be defined as the set of all mappings $\pi$ from the set of all algorithms to the three types of nodes$\{E,F,C\}$), which can be formulated as 
\begin{align}\label{eq:eq1}
\min&\quad\SMU(\A)=\sum_{i=1}^k\MU (R_i,\A),
\end{align}
but also to achieve that while maintaining equilibrium in memory usage,
\begin{align}\label{eq:eq2}
\min&\quad\sum_{i=1}^k\bigg(\MU (R_i,\A)-\frac{\SMU(\A)}{k}\bigg)^2.
\end{align}

\begin{figure}[tb]\centering
\includegraphics[width=0.4\linewidth]{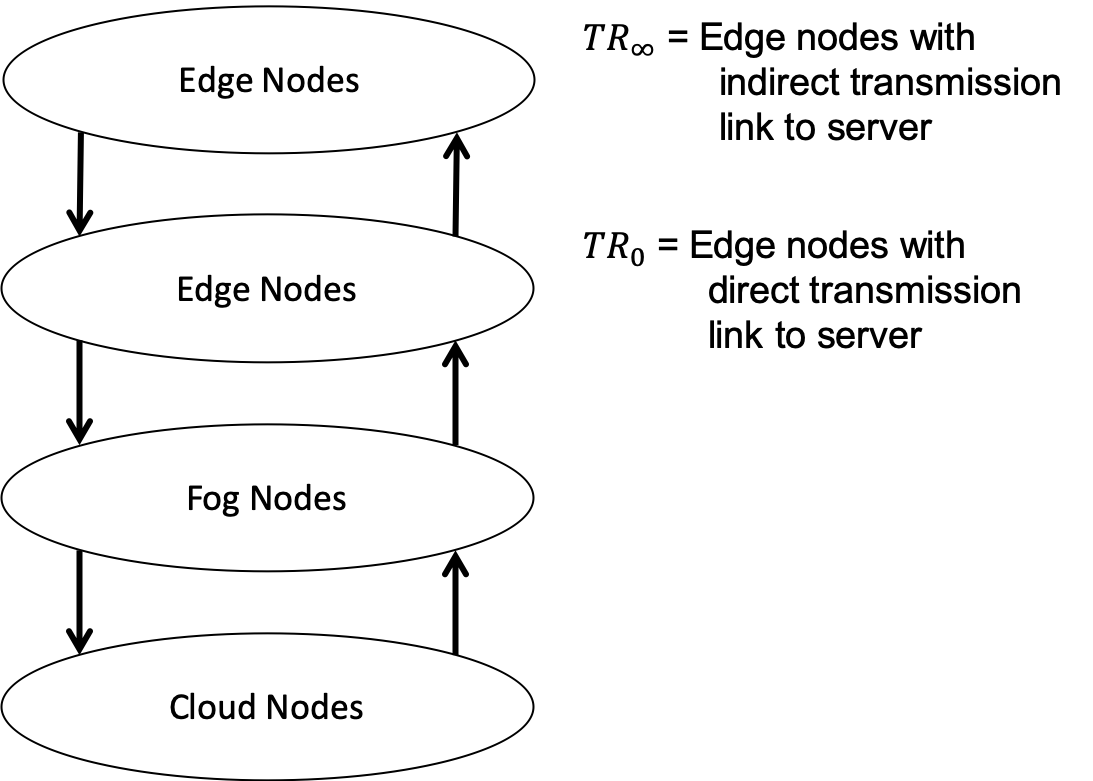}
\caption{Cooperative robotics network: edge nodes are divided into two classes $TR_0$ and $TR_{\infty}$, depending on the existence of a direct transmission link to a fog node.}
\label{fig1p}
\end{figure}

\begin{figure}[tb]\centering
\includegraphics[width=0.4\linewidth]{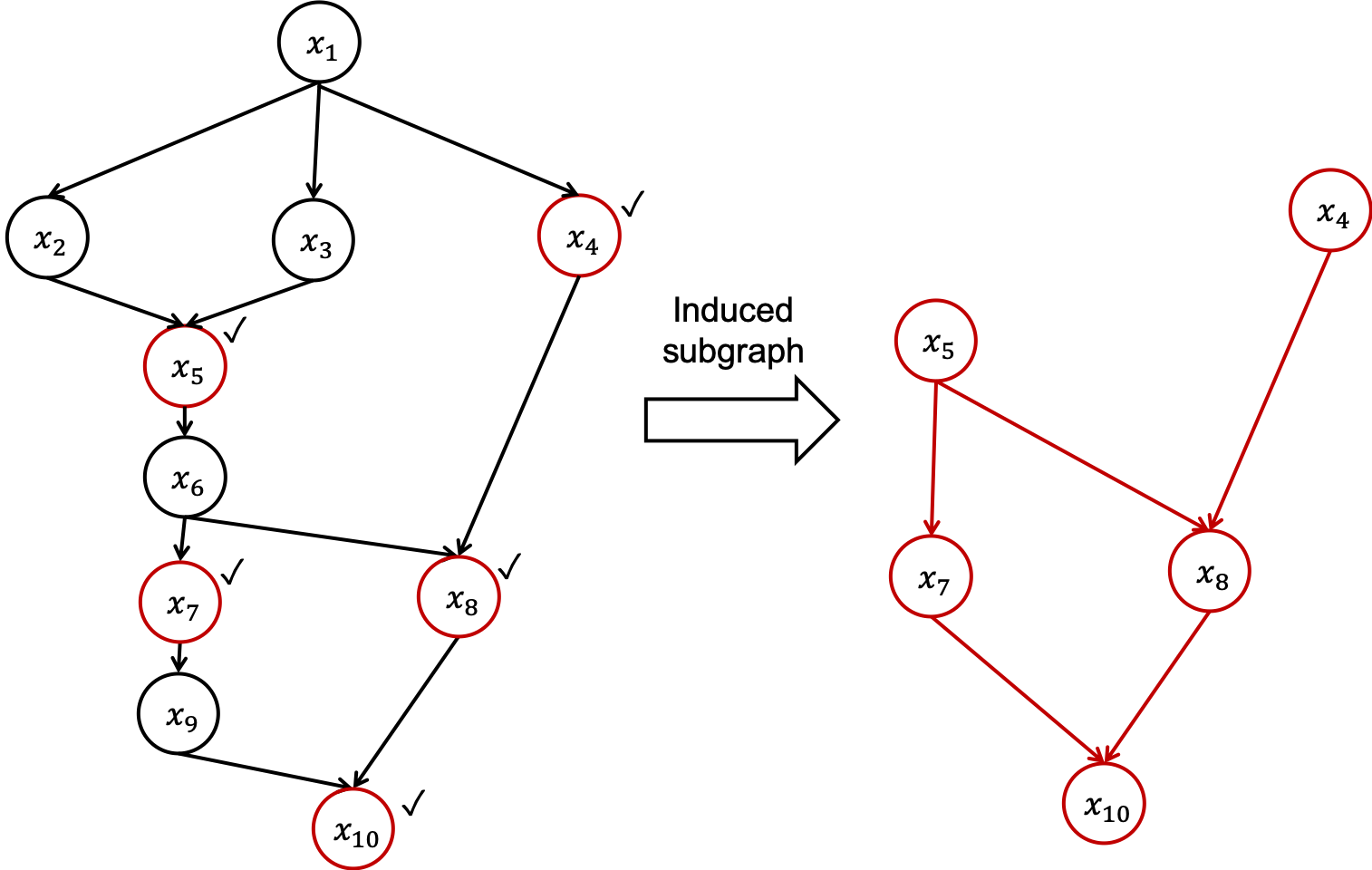}
\caption{Induced subgraph of a graph: the induced subgraph is considered with tick marked vertices.}
\label{figp}
\end{figure}

We can divide the edge nodes into two disjoint sets $TR_0$ and $TR_{\infty}$, see Figure~\ref{fig1p}:
\begin{itemize}
\item the set $TR_0$ contains all edge nodes with direct data transmission connection to a fog node;
\item the set $TR_{\infty}$ contains all edge nodes not included in $TR_0$, i.e., the set of all edge nodes that for transmitting some data to any fog nodes, the data necessarily needs to be transmitted through some other edge nodes.
\end{itemize}
\begin{example}
An example of $m_{in}(A_i)$ can be found in self-driving cars. The term $m_{in}(A_i)|_E$ can be seen in some cases, e.g., when road maps are the input and the presence of some obstacles due to accidents, roadblocks, etc., detected by some of the cars can be transmitted to other cars as additional input from the car that detect it in addition to the existing map. The term $m_{in}(A_i)|_{F\cup C}$ can be seen in some cases, such as when some new roads are made, some old roads are blocked permanently or temporarily due to changes in regulations, and this new information needs to be transmitted to all the cars as additional input from the fog or cloud in addition to the existing map. 
\end{example}
\begin{lemma}\label{lem:lem1}
If an algorithm $A_i$ is assumed to be allocated on an edge node, then the minimized memory usage by edge nodes in the statements \eqref{eq:eq1} and \eqref{eq:eq2} applies only to the case where the edge node is chosen from $TR_0$ or $m_{in}(A_i)|_{F\cup C}=0$.
\end{lemma}
\begin{proof}
First, assume that $m_{in}(A_i)|_{F\cup C}\neq0$. We will make a comparison between the two cases where the edge node chosen to execute algorithm $A_i$, $R_j$, comes from $TR_{\infty}$, and the case where $R_j$ comes from $TR_0$. Without loss of generality, we assume that the chosen edge node $R_j$ comes from $TR_{\infty}$ and is such that the total size $m_{in}(A_i)|_{F\cup C}$ of the data is transmitted from an edge node $R_{j'}$ in $TR_0$ to $R_j$. Some of the terms in the minimization statements \eqref{eq:eq1} and \eqref{eq:eq2} change with respect to the class containing $R_j$ by substituting the corresponding terms from Equation \eqref{eq:eq3}. The term $Const.$ refers to the unchanged subterm value in Equation \eqref{eq:eq3}. The following are the terms that will change in the minimization statements \eqref{eq:eq1} and \eqref{eq:eq2}:
\begin{itemize}
\item Suppose that $R_j$ is from $TR_0$. Then by Equation \eqref{eq:eq3}:
\begin{align}\label{eq:eq4}
\MU(R_j,\A)=Const.&+TM(R_j)+\left(m_{pr}(A_i)+m_{in}(A_i)|_{E}+m_{in}(A_i)|_{F\cup C}\right)
\end{align}
\item Suppose that $R_j$ is from $TR_{\infty}$. Then by Equation \eqref{eq:eq3}:
\begin{align}\label{eq:eq5}
\MU(R_j,\A)=Const.&+TM(R_j)+\left(m_{pr}(A_i)+m_{in}(A_i)|_{E}+m_{in}(A_i)|_{F\cup C}\right)
\end{align}
and
\begin{align}\label{eq:eq6}
\MU(R_{j'},\A)=Const.&+\left(TM(R_{j'})+m_{in}(A_i)|_{F\cup C}\right),
\end{align}
where the amount of $m_{in}(A_i)|_{F\cup C}$ must be transmitted to the edge node $R_j$ in addition to the original $TM(R_{j'})$.
\end{itemize}
By comparing the modified terms, it is easy to see that in the case where $R_j$ comes from $TR_{\infty}$, the total memory usage by all edge nodes is at least $m_{in}(A_i)|_{F\cup C}$ higher compared to the total memory usage by all edge nodes for the case where $R_j$ comes from $TR_0$. 
\end{proof}
Lemma~\ref{lem:lem1} implies that to obtain the minimum memory usage by all edge nodes, for an algorithm with additional necessary input from the cloud or fog, and if it must be executed on an edge node, that edge node must be in the class $TR_0$.
\begin{definition}
Let $G=(V,E)$ be a graph. Let $v_1,v_2,\ldots,v_n\in V$ be a sequence of vertices of the graph $G$. We say that $v_1\ldots v_n$ is a \emph{path} in $G$ if each two consecutive vertices in the graph $G$ are adjacent, i.e., there is an edge of $E$ connecting the two vertices.
\end{definition}
\begin{definition}
A subgraph $G'=(V',E')$ is called an \emph{induced subgraph} if $V'\subseteq V$ and two vertices $v_1$ and $v_2$ in $V'$ are connected by an edge in $E'$, if there is a path, $v_1w_1\ldots w_kv_2$, between them in $G$ such that 
$$\{w_1, \ldots, w_k\}\cap V'=\emptyset,$$ 
i.e., the intermediate vertices are not in $V'$. See Figure~\ref{figp}. 
\end{definition}

Note that, for a directed graph $G$, the path must be switched to the directed path, and the induced subgraph can be defined with respect to the directed path. For more details and properties, see \cite{bondy:2008}.

The next proposition states that there is a method for allocation of the algorithms with optimal memory with respect to the minimization statements \eqref{eq:eq1} and \eqref{eq:eq2}. The method is described in detail in the proof, which is based on recursive allocation and sorting, see \cite{Ours:2020imp}.

\begin{proposition}{(Balancing Method (\cite{Ours:2020imp}))}\label{prop:prop1}
Let $A=\{a_1, \ldots, a_n\}$ be a list of positive real numbers and $m\geq 1$ be an integer. We assume that we randomly remove one of the $a_i$'s and put it in exactly one of the $m$ places, and do the same process until the set $A$ is empty. Let $X_j$ for $j=1,\ldots,m$ be random variables defined as the sum of $a_i$'s at location $j$. Then there is an allocation such that the variance of $X_i$'s is minimal.

Moreover, for the case $A=\{a_1,\ldots,a_n\}$ and $m\geq n$, the minimum variance can be obtained by setting each $a_i$ in an empty position.
\end{proposition}
\begin{corollary}{(\cite{Ours:2020imp})}\label{cor:cor1}
In Proposition~\ref{prop:prop1}, we can assume that some of the $m$ places have some fixed initial values. Then we apply the method in the proof of Proposition~\ref{prop:prop1} for the allocation of the numbers in the list of numbers under the assumption that some of the $m$ places have initial values. The method under the assumption that some initial values exist does not give us the overall minimum variance with respect to the case without considering the initial values, but it does give us the optimal allocation such that the variance is minimum with respect to the existence of the initial values. 
\end{corollary}
Note that in a robotic network cloud system, to find a solution for the optimal memory usage in the allocation problem, we first need to find a set of solutions with minimum total memory usage by all edge nodes (or all fog nodes or all cloud nodes), then we need to solve the memory balancing problem. First, we take the set of all edge nodes as a single so-called universal edge, denoted by $\mathcal{E}$, similarly for the fog and the cloud.

Now we use the solution of the allocation problem for a cloud robotic system with a single robot proposed in \cite{Ours:2020}. This gives us a set of algorithms to run on the universal edge, the universal fog, and the universal cloud. That is, we already know which algorithms need to be executed on the edge nodes. 

Now we reason about the balancing problem. Suppose that the solution consists of 
$$A^e=\{A^e_1,\ldots,A^e_{a_1}\}\cup\{A^e_{a_1+1},\ldots,A^e_{a_2}\}\subset A,$$ 
as the set of algorithms to run on the edge nodes, such that 
$$\forall B\in \{A^e_1,\ldots,A^e_{a_1}\}, m_{in}(B)|_{F\cup C}\neq0,$$
and 
$$\forall B\in\{A^e_{a_1+1},\ldots,A^e_{a_2}\}, m_{in}(B)|_{F\cup C}=0.$$

Now we apply Proposition~\ref{prop:prop1} first to $\{A^e_{a_1+1},\ldots,A^e_{a_2}\}$ and to all edge nodes in class $TR_0$ by considering the list of values as the total memory needed for the input and processing of the algorithms, and the edge nodes as places. After completing this process, we again apply Proposition~\ref{prop:prop1} and Corollary~\ref{cor:cor1} for $\{A^e_1,\ldots,A^e_{a_1}\}$ and to all edge nodes, but with prior information about the memory usage by the edge nodes in class $TR_0$.

\begin{remark}\label{rem:rem1}
Note that if we decide to allocate several algorithms to an edge node, the total memory can be obtained by applying the algebra of memory, see \cite{Ours:2020}, to the induced subgraph consisting of all algorithms that need to be executed on the edge node.  
\end{remark}
\begin{remark}\label{rem:rem2}
Assume that the number of edge nodes is $k$. Similar to the evaluation of the time parameter, for any given algorithm allocation, the memory usage by the edge nodes can be represented as $(m_1,\ldots,m_k)$, where $m_i$ is the total memory usage by the edge node $i$ for a given algorithms allocation. Since we now assume that the edge nodes have no differences, we take the maximum value of the list for all edge nodes' required memory.
\end{remark}

\subsection{Memory-Time Optimization}
Considering the remarks~\ref{rem:rem0} and \ref{rem:rem2}, similar to \cite{Ours:2020}, the problem of optimizing the memory capacity with respect to the time parameter will be to find the list of solutions for the time component along with the list of solutions for the memory component and then depending on the component of interest, we can find the solution as the point in 2-D time-memory space that is closest to the origin.

\subsection{Method}
Our method can be explained as follows:
\begin{itemize}
\item[] \textbf{Input:} The following information is needed:
\begin{enumerate}
\item List of algorithms $\{A_1,\ldots,A_n\}$; 
\item Execution dependency of the algorithms; 
\item Execution time of algorithms on all nodes of cloud, fog, and edge; 
\item Data transmission time between all nodes of the cloud, fog, and edge;
\item Input, processing and output memory size of algorithms;
\item Additional information about algorithms: where to run them (optional, to account for known constraints);
\item Identify the classes of nodes on edge $E$, on cloud $C$, and on fog $F$.
\end{enumerate}
Note that, the values for points 3 and 4 can be estimates. The closer they are to the real values, the better the allocation will be.
\item[] \textbf{Output:} The mapping $\pi:\{A_1,\ldots,A_n\}\rightarrow\{E,F,C\}$ for algorithm allocation such that it gives the optimal overall time and memory. The mapping $\pi$ maps each algorithm to a corresponding node in either classes $E$, $F$, and $C$.
\item[] \textbf{Steps:} The following steps should be followed:
\begin{enumerate}
\item Construct the graph of algorithms,$G$, and its respective semi-lattice $\mathcal{ SL }(G)$; 
\item Find the set of all execution flows, $\mathrm{ExecutionFlows}(G)$;
\item Make a guess on the optimal allocation algorithms and find its overall time and memory;
\item Apply the branch and bound algorithm to the elements of $\mathrm{ExecutionFlows}(G)$, one at a time as follows. Note that allocation of $\mathbf{1}$ and $\mathbf{0}$ are on the same edge node, and for the other algorithms, it is on a subset of $\{E,F,C\}$ (the subset because with the prior information about the algorithms, we might be able to remove some nodes from at least one of the classes of $E$, $F$ and $C$. The number of branches is equal to the number of nodes in all classes):
\begin{enumerate}
\item Apply the algebra of time \cite{Ours:2020} to the subterms of $\mathrm{ExecutionFlows}(G)$ induced by considering the algorithms already allocated in the previous steps, and find the partial overall time. And apply the algebra of memory \cite{Ours:2020} to the subterms of $\mathrm{ExecutionFlows}(G)$ induced by considering algorithms already allocated in previous steps and find the partial overall memory.
\item Compare the results of the previous step with the guessed optimal solution. If the distance to the origin is higher, then stop the branching. And if it is less or equal, proceed to the next algorithm;
\item If all algorithms are allocated, and the distance to the origin is less than the guessed optimal solution, then update the guess on the optimal allocation algorithms with the current overall time and memory and proceed to the next possible branch.
\end{enumerate}
\item The updated guess on the optimal allocation after completing the previous steps is the optimal allocation.
\end{enumerate}
\item[] \textbf{Special Cases:} If we are interested in the optimal time or the optimal memory usage, then:
\begin{enumerate}
\item For optimal time: find the solutions to the problem statement~\eqref{optnew};
\item For optimal memory: apply the method in \cite{Ours:2020} considering universal nodes, apply Proposition~\ref{prop:prop1} and Corollary~\ref{cor:cor1}.
\end{enumerate}
\end{itemize}
The preceding method is similar to the method in \cite{Ours:2020} and the differences are that here the number of branches is the number of nodes and it is assumed to be at least $3$ (we consider architectures with at least $3$ nodes instead of with exactly $3$ nodes). The pseudocode of the algorithm is shown in Algorithm \ref{algpseud}. Also, when evaluating the time parameter, we first obtain several values that will be the set of time values for each edge node, taking into account their distances to the origin in the hyperspace of the edge nodes' time values. Moreover, for the particular case in which we want to find the optimal memory usage by robots, a simpler solution is formulated and described in the section memory optimization.

\begin{algorithm}
\caption{Optimal algorithm allocation minimizing time and memory}\label{algpseud}
\begin{algorithmic}[1]
\Procedure{}{$\mathrm{ExecutionFlows}(G),\mathrm{Archi.}(E,F,C)$}
\State $\mathrm{Map} \gets \{(A,x)\mid A\in G, x\in E\cup F\cup C\}$
\State Allocate all algorithms to a fog node $f\in F$ and evaulate $M \gets \textit{Memory(G,f)}$ and $T\gets \textit{Time(G,f)}$. \Comment{Use algebra of memory and time in \cite{Ours:2020}}
\State $h \gets \textit{max}\{\mathrm{height}(e)\mid e\in \mathrm{ExecutionFlows}(G)\}$. \Comment{$\mathrm{height}$ of a path is the number of vertices of the path.}
\For{$l:=h$ to $1$}
\State $\mathcal{A} \gets \{A\mid A \textit{ is a node of height $l$ of $e$}, e\in \mathrm{ExecutionFlows}(G) \}$.
\State $\mathrm{TempMap} \gets \{(A,x)\mid A\in \mathcal{A}, x\in E\cup F\cup C\}$ \Comment{Branching Algorithms.}
\For{$X\in \mathrm{TempMap}$}
\State $M_1 \gets \textit{PartialMemory($X$,$\mathrm{TempMap}$)}$ and $T_1\gets \textit{PartialTime($X$,$\mathrm{TempMap}$)}$ \Comment{Use algebra of memory and time in \cite{Ours:2020} to obtain partial memory usage and time for $X$ given the allocation $\mathrm{TempMap}$}
\If {$\sqrt{(M_1/M)^2+(T_1/T)^2}>\sqrt{2}$} \Comment{Bounding.}
\State $\mathrm{TempMap} \gets \mathrm{TempMap}\setminus\{X\}$
\EndIf
\EndFor
\State $\mathrm{Map}\gets \mathrm{Map}\cap\mathrm{TempMap}$
\EndFor
\State\Return $\mathrm{Map}$
\EndProcedure
\end{algorithmic}
\end{algorithm}

\subsection{Examples}
We will now present some examples that serve as a way to clarify the method's description. Further examples for independent explanation of the time and memory parameters can be found in the Supplemental material~\ref{sup}.

\begin{figure}[tb]\centering
\includegraphics[width=0.2\linewidth]{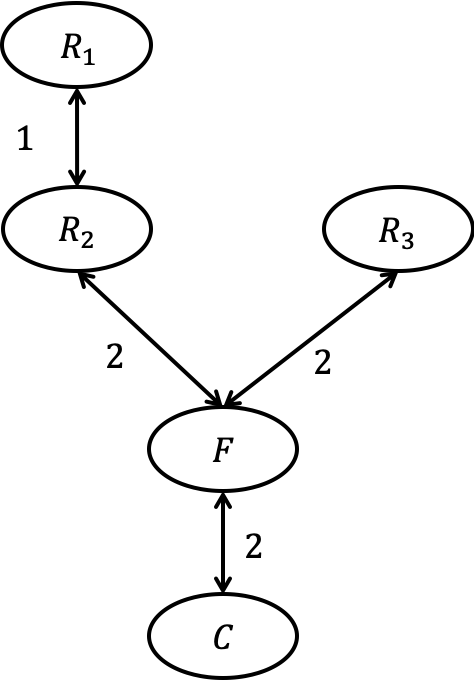}
\caption{Three robots cloud system. The numbers on the edges are the average data transmission times in seconds.}
\label{figr}
\end{figure}

\begin{figure}[tb]\centering
\includegraphics[width=0.2\linewidth]{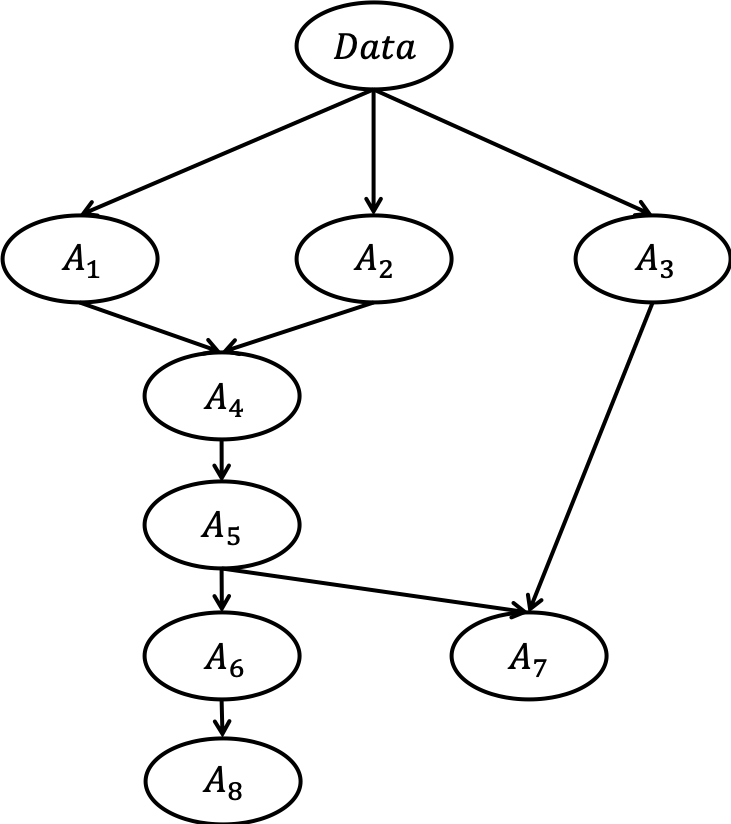}
\caption{The graph of $9$ algorithms.} 
\label{figa}
\end{figure}

\begin{comment}
\subsection{Memory-Time}
\end{comment}
We performed a test for three robots, with architecture and the data transmission times, shown in Figure~\ref{figr}. We use $9$ algorithms, and the graph of the algorithms is shown in Figure~\ref{figa}.

For the memory usage of the dataset, we assume the maximum size of $500$ $Mbytes$, the algorithms $A_1,\ldots,A_8$ use respectively $300$, $50$, $100$, $200$, $100$, $150$, $50$, and $200$ $Mbytes$, and also with different space complexities respectively $O(n^2)$, $O(n)$, $O(\log(n))$, $O(n\log(n))$, $O(n)$, $O(log(n))$, $O(n^2)$, and $O(n \log(n))$. The average execution time of the algorithms, depends on the location where they are executed, as given in Table~\ref{tabexec}. We also assume that the robots are identical in terms of processing power.

\begin{table}[tb]
\caption{The average execution time of algorithm nodes in cloud system's nodes (in seconds). The average execution time of Data is assumed to be $0$.}
\begin{center}
\begin{tabular}{l|cccccccc}
&$A_1$&$A_2$&$A_3$&$A_4$&$A_5$&$A_6$&$A_7$&$A_8$\\
\hline
Edge$(R_1)$&2&4&6&2&4&6&2&4\\
Edge$(R_2)$&2&4&6&2&4&6&2&4\\
Edge$(R_3)$&2&4&6&2&4&6&2&4\\
Fog$(F)$&1&2&3&1&2&3&1&2\\
Cloud$(C)$&0.5&1&1.5&0.5&1&1.5&0.5&1\\
\end{tabular}
\end{center}
\label{tabexec}
\end{table}

\begin{figure}[tb]\centering
\includegraphics[width=0.5\linewidth]{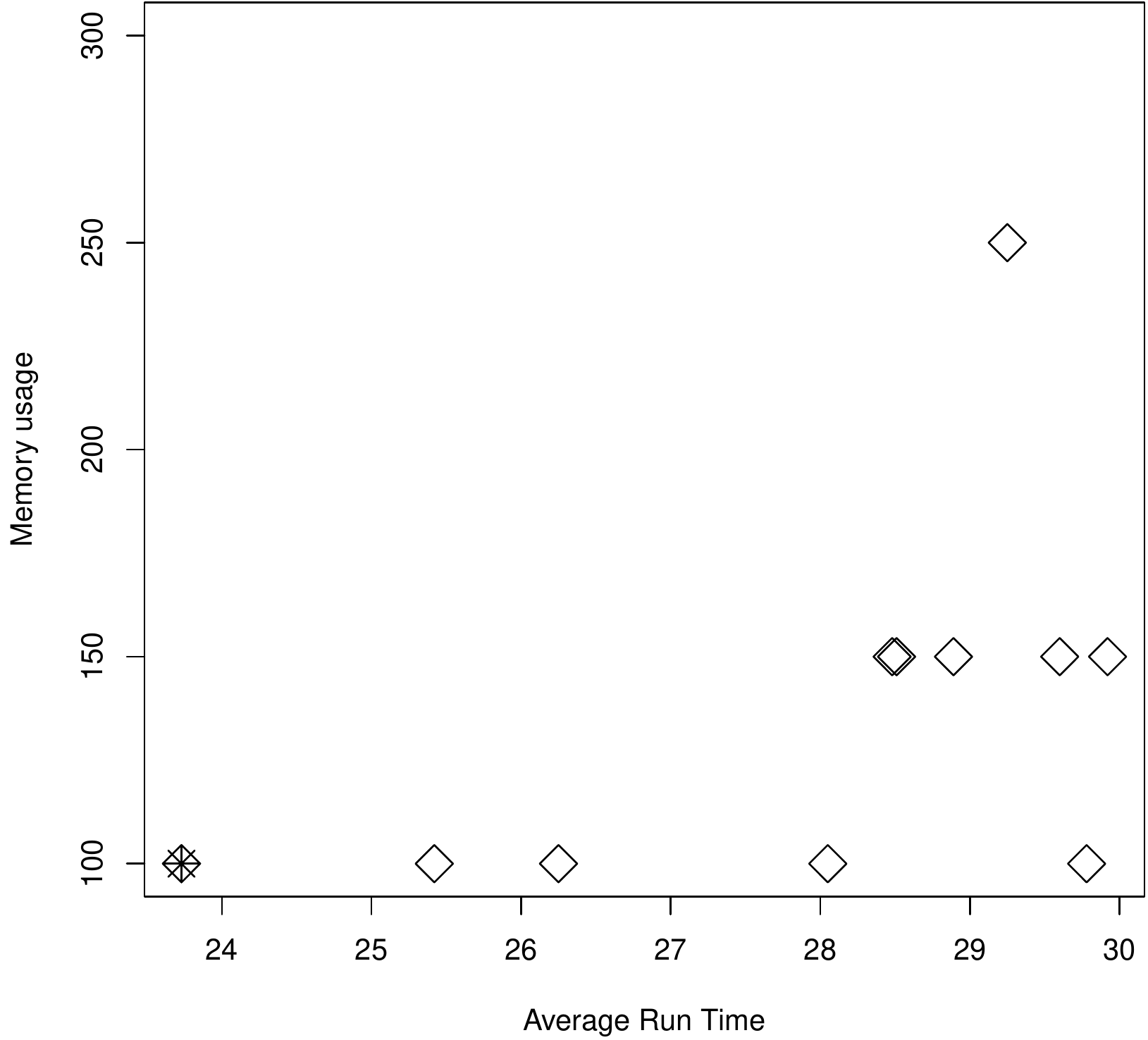}
\caption{Identifying the minimum Memory-Time values, the average run time is in seconds and the memory usage is in $Mbytes$. The star correspond to the point with minimum Memory-Time value. The diamonds are all possible Memory-Time solutions. We only show the points where the maximum memory usage by robots is less than $300 Mbytes$ and the overall completion time is less than $30$ seconds.} 
\label{figs}
\end{figure}

The result of applying the proposed method to optimize the Memory-Time, can be seen in Figure~\ref{figs}. The optimal solutions are the points closest to the lower left corner (origin). Next, we multiply each of the data transmission time in Figure~\ref{figr} by $5$, i.e., all direct data transmissions between nodes will be $5$ times slower than shown in Figure~\ref{figr}. Finally, we multiply each of the data transmission time in Figure~\ref{figr} by $0.1$, i.e., all direct data transmissions between nodes will be $10$ times faster than shown in Figure~\ref{figr}. The values $5$ and $0.1$ were chosen to emphasize the importance of the data transmission time, so that for $0.1$, the average execution times of all algorithms are greater than the data transmission time, and for $5$ the average execution times of most algorithms are smaller than the data transmission time. In Table~\ref{tabsall}, we show the algorithm's allocations that yields the optimal solutions and minimizes the Memory-Time for all three scenarios.

\begin{table*}[tb]
\caption{The algorithms allocation that minimize the Memory-Time. $Exp. 1$ is with data transmission times shown in Figure~\ref{figr}. $Exp. 2$ is with data transmission 5 times slower than the times shown in Figure~\ref{figr}. $Exp. 3$ is with data transmission 10 times faster than the times shown in Figure~\ref{figr}.}
\begin{center}
\resizebox{\textwidth}{!}{
\begin{tabular}{rccccccc}
&Average Time&Maximum Memory&&&&&\\
&(in seconds)&Usage by Robots&&Edge&&Fog&Cloud\\\cline{4-6}
&&(in $Mbytes$)&Robot 1&Robot 2&Robot 3&&\\
\hline
~~$\vline$&\textbf{23.73}&\textbf{100}&-&-&-&-&All\\
Exp. 1~~$\vline$&\textbf{23.73}&\textbf{100}&-&-&-&$A_7$&Data, $A_1,\ldots,A_6$, and $A_8$\\
~~$\vline$&\textbf{23.73}&\textbf{100}&-&-&-&$A_3$ and $A_7$&Data, $A_1, A_2, A_4, A_5, A_6$, and $A_8$\\
\hline
~~$\vline$&\textbf{58.31}&\textbf{100}&-&-&-&All&-\\
Exp. 2~~$\vline$&84.11&100&-&-&-&-&All\\
~~$\vline$&84.11&100&-&-&-&$A_7$&Data, $A_1,\ldots,A_6$, and $A_8$\\
~~$\vline$&84.11&100&-&-&-&$A_3$ and $A_7$&Data, $A_1, A_2, A_4, A_5, A_6$, and $A_8$\\
\hline
~~$\vline$&\textbf{10.16}&\textbf{100}&-&-&-&-&All\\
~~$\vline$&\textbf{10.16}&\textbf{100}&-&-&-&$A_3$&Data, $A_4,\ldots,A_8$, $A_1$, and $A_2$\\
Exp. 3~~$\vline$&\textbf{10.16}&\textbf{100}&-&-&-&$A_7$&Data, $A_1,\ldots,A_6$, and $A_8$\\
~~$\vline$&\textbf{10.16}&\textbf{100}&-&-&-&$A_3$ and $A_7$&Data, $A_1, A_2, A_4, A_5, A_6$, and $A_8$\\
~~$\vline$&10.34&150&-&$A_7$&-&-&Data, $A_1,\ldots,A_6$, and $A_8$
\end{tabular}
}
\end{center}
\label{tabsall}
\end{table*}

Minimizing the overall time is as important as minimizing the memory, and it is not enough to minimize one and describe the performance of the system using only that information. By focusing only on minimizing the memory usage by all robots, we avoid the overall execution time, and thus the system can be more time-consuming. And by focusing only on minimizing the time, at least one of the robots may not have enough memory and thus may not be able to execute some algorithms. 

As refered in the problem statement~\eqref{optnew}, finding the solutions is equivalent to finding the optimal time parameter of the algorithms' allocation for a single robot cloud system in \cite{Ours:2020}.

\subsection{Edge Node Settings}
Throughout the manuscript, we assumed that all the edge nodes are identical, and none has an advantage over the others. If we drop this assumption, the set of all edge nodes, $E$, can be viewed as a union of finitely many disjoint subsets, where each subset contains all identical edge nodes. That is $E=\bigcup_{i=1}^e E_i$, where for $j\neq i$, $E_i\cap E_j=\emptyset$, the elements of $E_i$'s are identical edge nodes, and $e$ is minimal, i.e., if we find $E=\bigcup_{i=1}^{e'} E'_i$, then $e'\geq e$. Let $A(E_i)$ be the set of all algorithms that will be used in the edge nodes of $E_i$. Note that, here we only need to balance the memory usage of the edge nodes in each $E_i$'s. 

Now, to solve the allocation problem, for each $i=1, \ldots, e$, apply the provided method for the induced subgraph of algorithms considering all the algorithms occurring only in the set $A(E_i)$, and the induced subgraph of the cloud robotics architecture omitting all the edge nodes except $E_i$. Then, for each $i,j=1, \ldots, e$ with $i\neq j$, apply the proposed method for the induced subgraph of algorithms considering all the algorithms that occurred only in the set $A(E_i)\cap A(E_j)$ but were not allocated before, and the induced subgraph of the cloud robotics architecture dropping all the edge nodes except the $E_i\cup E_j$. Iterate the same arguments for more than three sets, until all algorithms are allocated.

\subsection{Arbitrary Number of Fog, Cloud and Intermediate Nodes}
We can classify the nodes of a cloud robotics architecture into two classes: all edge nodes and all other nodes. The other nodes can include fog nodes, cloud nodes, dew nodes (on-premises smart things that are independent of the cloud but in collaboration \cite{botta:2019}), and any other nodes with processing capabilities that are physically separate from robots. Our method provides ways to minimize the maximum memory usage by all the robots and to minimize the maximum overall time, so that all the robots have outputs of all the algorithms while minimizing both memory usage and time. In the calculations, the number of edge nodes does not change, the algebras of memory and time depend only on the graph of the algorithms, and the calculations are independent of the classifications of other nodes. Therefore, the same methods work perfectly if we change the classification names of all the nodes except the edge nodes. In this case, we only need new partition names for all the nodes.

Our method also works when we consider multiple fog nodes and cloud nodes. Moreover, if we consider a direct communication link between edge nodes and cloud nodes, we only need to combine fog nodes and cloud nodes in Figure~\ref{fig1p} and define $TR_0$ and $TR_{\infty}$ as the set of edge nodes with direct and indirect transmission links to the cloud and server, respectively. Then, our method also works in this case to find the optimal static algorithm allocation that minimizes memory and time.

\section{Experiments}
We have collected data from the real-world and use it as input parameters in simulations. We first describe the data and then apply our method to the simulations with random architectures. After collecting the data, the experiments are performed on a HP Laptop 15-dw2xxx with Intel Core i5 10th generation with processor Intel(R) Core(TM) i5-1035G1 CPU @ 1.19 GHz, RAM 16.0 GB, 64-bit operating system, and we used RStudio Version 1.4.1103 © 2009-2021 RStudio, and the R version 4.0.3 (2020-10-10) copyright © 2020 The R Foundation for Statistical Computing Platform. The code is made available at \url{https://github.com/SaeidZadeh/StaticAllocation}.
\subsection{Data}
Each edge node is a Raspberry Pi 4 Model B. The fog node is a computer with 32GB of RAM, Intel(R) Core(TM) i7-9700K CPU @ 3.60GHz with 8 cores and an Nvidia GeForce RTX 2080 GPU. The cloud node is a virtual machine with 64GB of RAM, Intel(R) Xeon(R) Silver 4114 CPU @ 2.20GHz with 2 cores and an Nvidia Tesla V100 PCIe 16GB GPU.

Algorithm $A_1$ reads a folder of images of a person's face. The algorithm extracts features from each image and creates a database that associates a person with the features of their face. The second algorithm, $A_2$, stores the database created by its predecessor in a file on disk. $A_3$ loads the database file back into memory so that operations can be performed on it. $A_4$ receives a compressed image of a person's face and decompresses it. $A_5$ extracts facial features from an input image. $A_6$ compares the features of a single image with all images in the database. $A_7$ receives the result of the matching algorithm and identifies the person by finding the image with the most matches.

The different nodes are all different machines within the same network and have fixed IP addresses. They are connected to the network through a router via Ethernet cables. They communicate with each other using the node's local IP address through sockets that use the TCP/IP protocol. To capture the execution times of an algorithm on an executing node, we obtain the Unix timestamps immediately before and after the execution of the algorithm on the node. The execution time of the algorithm on each node is the difference between the two timestamps. The average execution time of an algorithm on an executing node is the average execution time of the algorithm on the node obtained for 100 trials.

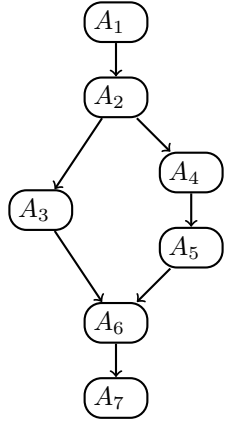
\begin{figure}[tb]\centering
\begin{tikzpicture}
\begin{scope}%[every node/.style={rectangle,thick,draw,rounded corners=.8ex}]
    \node (A) at (0,0) [text width=0.6cm,rectangle,thick,draw,rounded corners=1.5ex] {$A_1$};
    \node (B) at (0,-1)  [text width=0.6cm,rectangle,thick,draw,rounded corners=1.5ex] {$A_2$};
    \node (C) at (-1,-2.5)  [text width=0.6cm,rectangle,thick,draw,rounded corners=1.5ex] {$A_3$};
    \node (D) at (1,-2) [text width=0.6cm,rectangle,thick,draw,rounded corners=1.5ex]{$A_4$};
    \node (E) at (1,-3)[text width=0.6cm,rectangle,thick,draw,rounded corners=1.5ex] {$A_5$};
    \node (F) at (0,-4) [text width=0.6cm,rectangle,thick,draw,rounded corners=1.5ex]{$A_6$};
    \node (G) at (0,-5)[text width=0.6cm,rectangle,thick,draw,rounded corners=1.5ex] {$A_7$};

\end{scope}

    \path [->] (A) edge[thick,->] (B);
    \path [->] (B) edge[thick,->] (C);
    \path [->](B) edge [thick,->] (D);
    \path [->](C) edge [thick,->] (F);
    \path [->](D) edge [thick,->] (E);
    \path [->](E) edge [thick,->] (F);
    \path [->](F) edge [thick,->] (G);
\end{tikzpicture}
\caption{Graph of all algorithms}
\label{fig1pp}
\end{figure}

The graph of all algorithms is shown in Figure~\ref{fig1pp}.
Table~\ref{tab2p} shows the communication time between nodes for transferring $32$ bytes of data.

\begin{table}[tb]
\caption{Transmission time of $32$ bytes of data, in seconds. $\mathcal{ FN }(\mu,\sigma)$ is used to denote the folded normal distribution with parameter $\mu$ and standard deviation $\sigma$. $F$ is used to denote the fog, $C$ is used for the cloud, and $E$ is used to represent any edge node.}\label{tab2p}
\begin{center}
\begin{tabular}{l|l}
Transmission&Time (in seconds)\\
\hline
$C\rightarrow F$&$0.439+\mathcal{FN}(0.109,0.087)$\\
$F\rightarrow C$&$0.417+\mathcal{FN}(0.376,0.365)$\\
$F\rightarrow E$&$0.475+\mathcal{FN}(0.187,0.397)$\\
$E\rightarrow F$&$0.447+\mathcal{FN}(0.182,0.111)$\\
$E\rightarrow E$&$0.112+\mathcal{FN}(0.061,0.023)$
\end{tabular}
\end{center}
\end{table}

Table~\ref{tab3p} shows all algorithms' average execution times to be executed by an edge node, fog, or cloud. Also, the input, output, and processing memory and space complexities of all the algorithms are shown.

\begin{table*}[tb]
\caption{Average execution time of algorithms (in seconds) on an edge node, fog, or cloud, along with their input (in bits), output (in bits), and processing memorys (in bytes), and their space complexities. $n$ is the number of images and $m$ is the size of image.}\label{tab3p}
\begin{center}
\begin{tabular}{l|ccccccc}
&$Edge$ (s)&$Fog$ (s)&$Cloud$ (s)&Input&Output&Processing&Space\\
&&&&Size (bits)&Size (bits)&Size (bytes)&Complexity\\
\hline
$A_1$&$0.445$&$0.153$&$0.047$&$4718592$&$1120$&$14619367$&$O(nm)$\\
$A_2$&$4.475$&$1.538$&$0.470$&$47185920$&$11200$&$11683901$&$O(n)$\\
$A_3$&$7.2\times10^{-4}$&$4.1\times10^{-4}$&$1.5\times10^{-4}$&$11200$&$11200$&$11684220$&$O(n)$\\
$A_4$&$2.0\times10^{-4}$&$7.74\times10^{-5}$&$3.46\times10^{-5}$&$11200$&$0$&$7799083$&$O(m)$\\
$A_5$&$6.61\times10^{-5}$&$1.94\times10^{-5}$&$9.96\times10^{-6}$&$11200$&$11200$&$11253700$&$O(m)$\\
$A_6$&$2.1\times10^{-4}$&$1.3\times10^{-4}$&$4.75\times10^{-5}$&$11200$&$1120$&$11261700$&$O(nm)$\\
$A_7$&$1.09\times10^{-3}$&$4.01\times10^{-3}$&$2.7\times10^{-4}$&$4718592$&$4718592$&$8010779$&$O(n)$
\end{tabular}
\end{center}
\end{table*}

We use these data as input parameters for simulated architectures and use our method to find the optimal algorihtm allocation and collect the results, and then compare them with the results using the proposed method in \cite{li:2018} for the same simulated architectures.
\subsection{Simulations}
The graph of algorithms  used in these experiments is shown in Figure~\ref{fig1pp}. For the architecture, we generated several random graphs. For $n$ edge nodes, $n=1,\ldots,10$, the architecture has $n+2$ nodes (2 is for the single cloud and single fog nodes), an edge (communication) from the cloud to the fog node, and from the fog to at least one of the $n$ edge nodes. To generate a random graph, we used Erdos-Renyi random graph generators, \cite{Erdos:1960}. Since the architecture must correspond to a connected graph, we need at least $n-1$ randomly placed edges between nodes. After each placement, we need to check whether the generated graph is connected or not. Then, we consider the generated graph, $G^1_n$, as the architecture of the network. The number of edges is randomly chosen from $[n-1,\frac{n(n-1)}{2}]$. After obtaining the generated graph, we search for the optimal allocation of algorithms, find the values of the average overall times required to transmit all outputs of all algorithms to all edge nodes, and then find the distance to the origin of all these values, denoted by $T(G^1_n)$.

Note that for the case $n=1$, there is only one possible valid architecture that we need to consider. For $n=2$, there are $4$ valid architectures, two of which are isomorphisms. To avoid graph repetition, we tested graph isomorphisms, \cite{Babai:1980}. To avoid architectural isomorphisms, we need to test the existense of permutations between adjacency matrices of all generated graphs of architectures.

To find out whether two graphs are isomorphic or not, we should check all permutations. To speed-up the test, we can perform a naive isomorphism test. That is, we find the sorted list of out-degrees and in-degrees of all nodes; if they are different, then the two graphs are not isomorphic. Note that the converse case does not hold.

\begin{comment}
\begin{figure}[tb]\centering
\includegraphics[width=1\linewidth]{PP1.png}
\caption{Example of non-isomorphic graphs whose sorted lists of out-degrees and in-degrees of all nodes are, however, identical.}
\label{fig0p}
\end{figure}
\end{comment}
For simplicity, for the case where the architecture has $n$ edge nodes, we considered and generated at most $n+5$ random non-isomorphic graphs. Note that in the case that after $5$ unsuccessful attempts to find a random non-isomorphic architecture, the next randomly generated architecture is used, regardless of whether it is isomorphic to the existing ones. This allows us to test the method with different architectures, if possible.

\begin{figure}[tb]\centering
\includegraphics[width=0.5\linewidth]{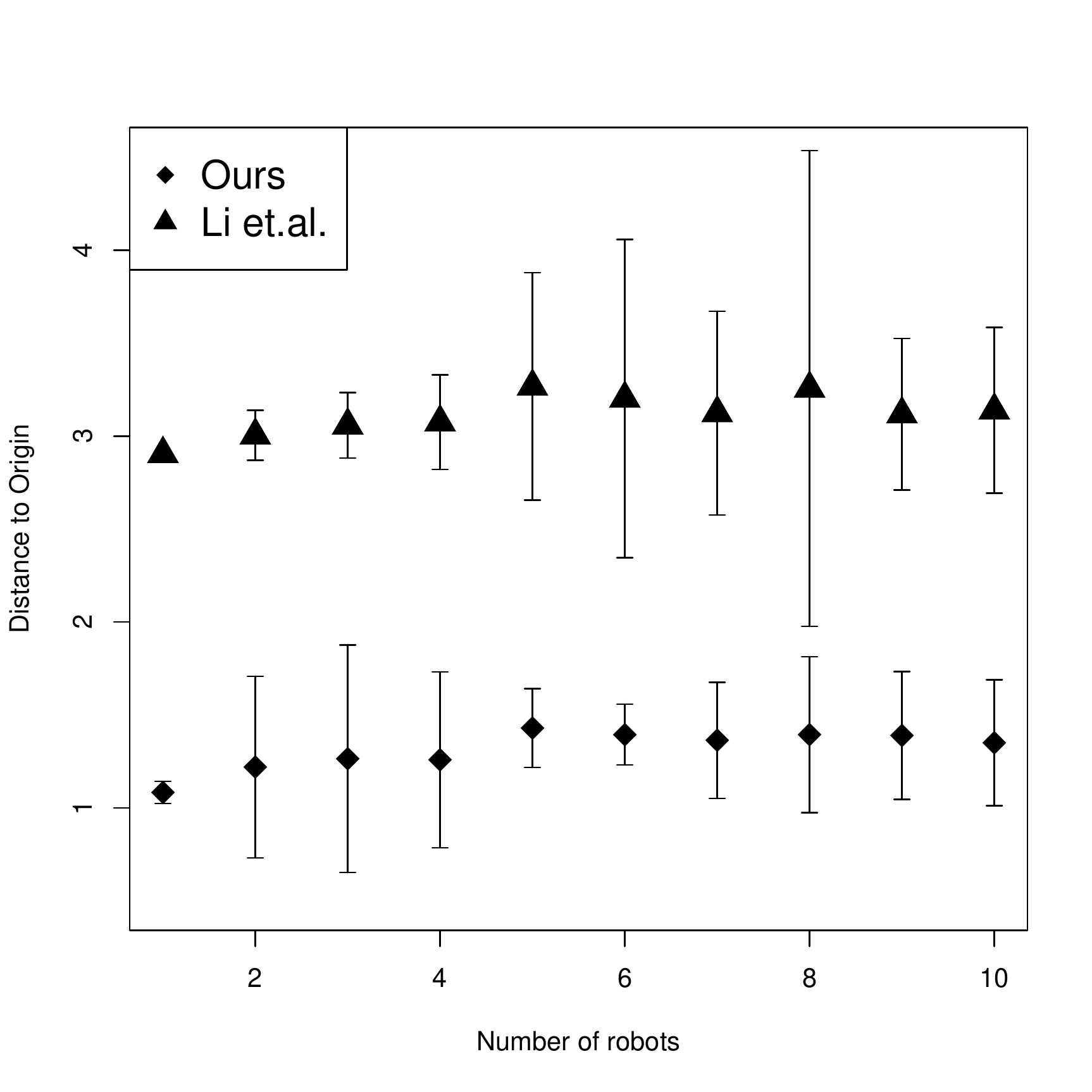}
\caption{Comparison of the distance to the origin for optimal algorithm allocation using our method, diamonds, and the method proposed in \cite{li:2018}, triangles, for randomly generated architectures with $n=1,\ldots,10$ robots. The bars are $99.9\%$ confidence intervals. The exact values can be found in the supplemental material~\ref{sup}, Table~\ref{tab4}.}
\label{figres}
\end{figure}

For each $n$, we generate random graphs, $G_n^2,\ldots,G_n^k$, multiple times. We applied our proposed method to simultaneously minimize the overall time and the maximum memory usage by the edge nodes. As described, the minimum value will be the shortest distance to the origin, which is computed by
\begin{equation}\label{eq:opt}
\sqrt{(t^{\pi}_1)^2+\ldots+(t^{\pi}_n)^2+(m^{\pi})^2}
\end{equation}
for each algorithm allocation $\pi$, where $\pi:\{A_1,\ldots,A_7\}\rightarrow\{E_1,\ldots,E_n,F,C\}$ is a mapping from all the algorithms to the set of all nodes of the architecture, where $t^{\pi}_i$ is the overall time required to transmit all outputs of all algorithms to the edge node $E_i$ when the algorithms are allocated to the nodes of the architecture according to the mapping $\pi$, and $m^{\pi}$ is the maximum overall memory required for the edge node with respect to the allocation mapping $\pi$ such that all robots have enough memory to process all algorithms assigned to them and store the outputs of all algorithms. We applied the method proposed by \cite{li:2018}, and at the same time, we applied our method to find the algorithm allocation. We compute the optimal solution of \eqref{eq:opt} using these two methods. For each randomly selected architecture, we find the optimal values for \eqref{eq:opt} using both methods, and we repeat this step $10$ times (this is to highlight the communication instability). We will choose the average of the obtained values as the solutions of both methods for the randomly generated architecture. To show that the performance of our method is independent of the architecture, for the case where we use $n$, $n=3,\ldots,10$, edge nodes, we randomly generate $n+5$ non-isomorphic architectures and find the optimal values for \eqref{eq:opt} using both methods. Comparing the results obtained when considering $n=1,\ldots,10$ robots, we find that our method outperforms the method proposed in \cite{li:2018}. The result is shown in Figure~\ref{figres}. Note that after generating the first communication time with random delay for a randomly generated architecture, we let the bound on the distance to the origin using the branch and bound algorithm be the distance to the origin resulting from the allocation of all algorithms to the cloud, which is equal to $\sqrt{2}$.

Our method has two advantages over the method proposed in \cite{li:2018}: it simultaneously minimizes time and memory, and fully accounts for the communication time, i.e., the communication time to transmit the result of each algorithm to the node that initiated the request is included. As shown in \cite{Ours:2020}, there are certain intervals where the overall time for the optimal algorithm allocation differs from the optimal algorithm allocation proposed in \cite{li:2018}. These intervals illustrate the dependence of the optimal allocation on the communication time, which must be fully considered in the formulation. We fully incorporate the communication time in our method to find the true optimal results.

\begin{figure}[tb]\centering
\includegraphics[width=0.5\linewidth]{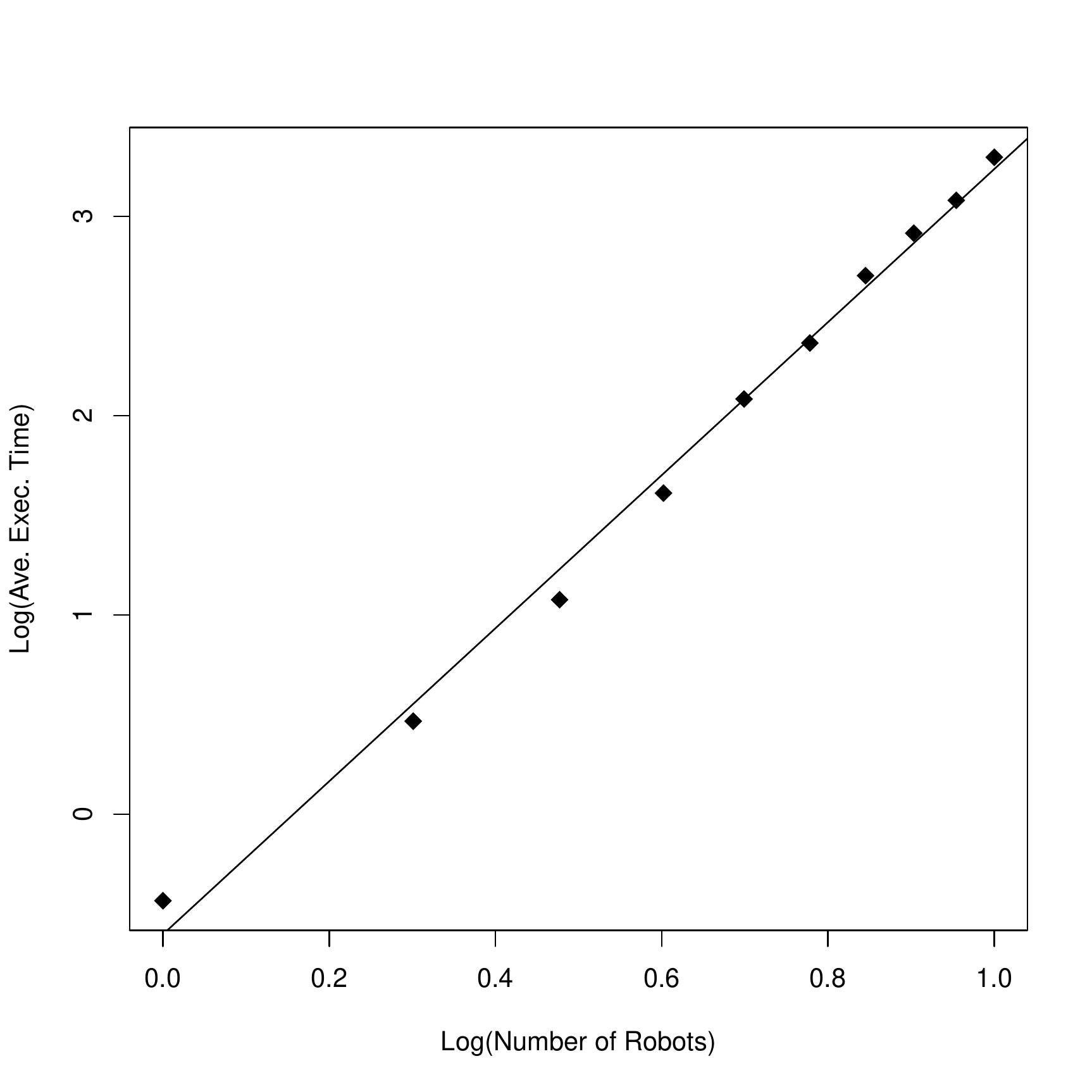}
\caption{The average execution time (in seconds) of the proposed algorithm, for randomly generated architectures with $n=1,\ldots,10$ robots, as a function of the number of robots, in logarithmic scale (all axis). Black diamonds are the real values and the line is the fit obtained by the linear regression with the $R^2=0.9941$ and $p$-value$=3.24e-10$. The exact values can be found in the supplemental material~\ref{sup}, Table~\ref{tabtime}.}
\label{figtime}
\end{figure}

The average algorithm execution time to obtain the results in Figure \ref{figres}, for randomly generated architectures with $n=1,\ldots,10$ robots, are shown in Figure \ref{figtime}.

\begin{comment}
\begin{center}
\begin{tabular}{llllllll}
Number&Average distance&Average distance&SD distance&SD distance\\
of robots&to the origin (Ours)&to the origin (\cite{li:2018})&to the origin (Ours)&to the origin (\cite{li:2018}) \\
\hline
1&$\mathbf{1.082624e+00}$&$2.904152e+00$&$1.797988e-02$&$1.337752e-05$\\
2&$\mathbf{1.21923998}$&$3.00364685$&$0.14804613$&$0.04088246$\\
3&$\mathbf{1.26423893}$&$3.05785760$&$0.18557013$&$0.05335121$\\
4&$\mathbf{1.25834091}$&$3.07446465$&$0.14334954$&$0.07702568$\\
5&$\mathbf{1.42877333}$&$3.26694388$&$0.06419105$&$0.18556065$\\
6&$\mathbf{1.39354458}$&$3.20185906$&$0.04949817$&$0.25943928$\\
7&$\mathbf{1.36342160}$&$3.12360627$&$0.09470016$&$0.16614295$\\
8&$\mathbf{1.3936499}$&$3.2559008$&$0.1269661$&$0.3878917$\\
9&$\mathbf{1.3889773}$&$3.1177981$&$0.1044364$&$0.1234059$\\
10&$\mathbf{1.349806}$&$3.13842$&$0.102876$&$0.1350024$
\end{tabular}
\end{center}
\end{comment}

\section{Scalability Analysis}
For a given architecture, like other algorithms searching for the longest path in a graph, our method's time complexity is in NP.

\begin{figure}[tb]\centering
\includegraphics[width=0.5\linewidth]{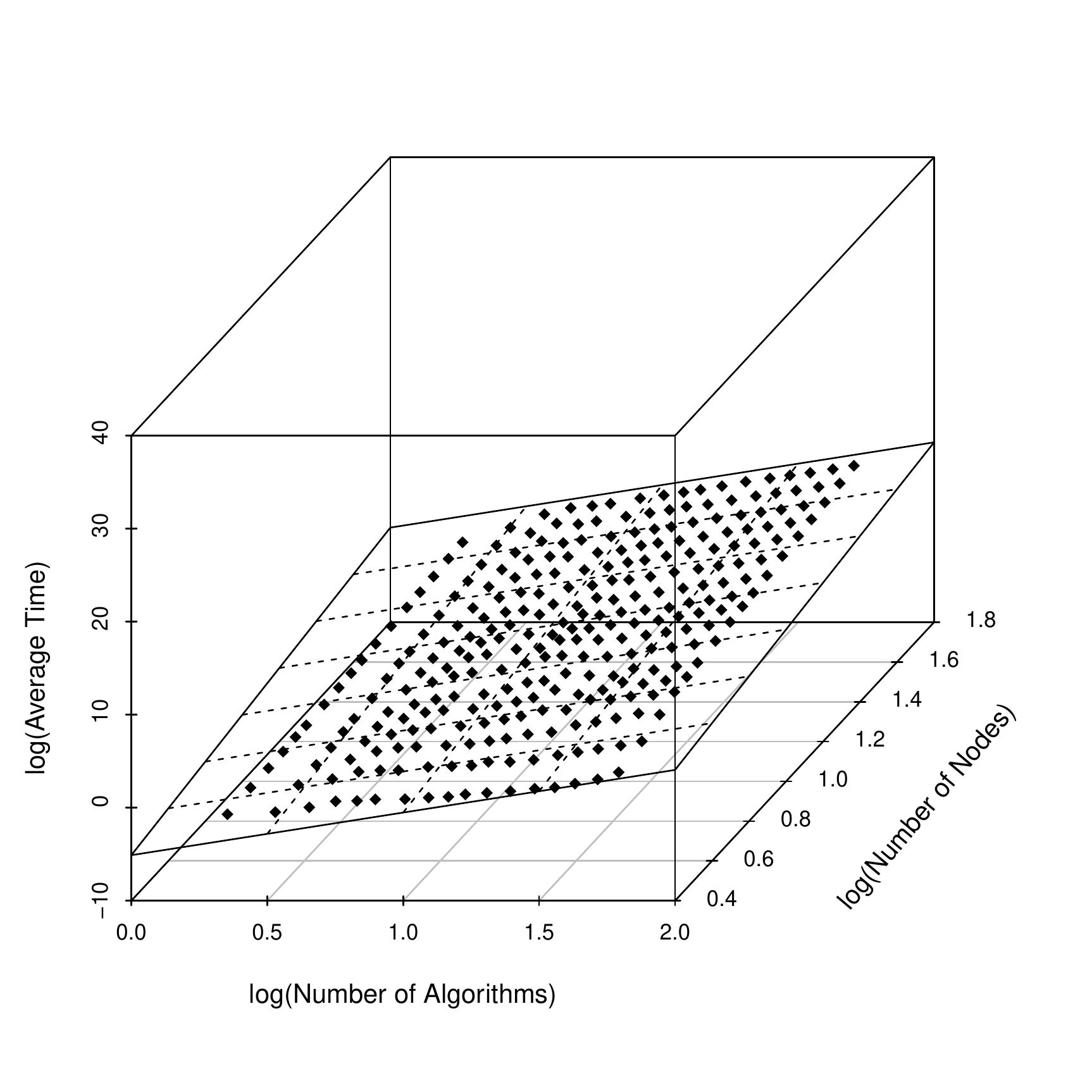}
\caption{Evaluation of scalability. Average time (in seconds) to find an optimal allocation as a function of the number of algorithms and as a function of the number of nodes, in logarithmic scale (all axis). Black diamonds are the real values and the plane is the fit obtained by the linear regression with the$R^2=0.9625$.}
\label{fig}
\end{figure}

To evaluate the scalability of our method, we conducted an experiment where we randomly generated the graph of all algorithms and randomly generated the architecture for a given number of nodes. We generate 10 random architectures and then set the number of algorithms and randomly generate 10 graphs of algorithms for each architecture. We plot the average time to find a solution for 10 graphs of algorithms in Figure \ref{fig}. The figure has all axis in logarithmic scale and shows a linear relationship between the average time and the number of algorithms to allocate, and also shows a linear relationship between the average time and the number of nodes, hence the time complexity of our method is polynomial.

\section{Conclusion}
We have provided a model that allows users, once they have decided on a particular architecture, to decide which type of robot is best suited to perform the desired tasks in terms of memory and computation capabilities, and also how algorithms should be allocated to achieve the fastest overall performance.

First, we minimized the average of the maximum overall time that all robots take between sending requests for any algorithm and receiving the response. Then, we minimized the total memory usage by all robots, so that all the robots have a nearly balanced memory usage. Since the minimum time and minimum memory do not necessarily hold for the same allocation of algorithms, we determine lower bounds such that the allocation problem for memory and time intersects with respect to these bounds. If the intersection has more than one solution to the allocation problem, the decision to assign the algorithms can be made by the user depending on the importance of the factors, memory usage, or overall time.

Our method provides a solution for achieving optimal performance of a robotic network cloud system and also allows comparison between the performances of multiple robot networks.

It works for edge nodes with different capabilities and for an arbitrary number of fog and cloud nodes.

The experimental results show that our proposed method outperforms \cite{li:2018}’s method with real data in minimizing the average execution time. The time optimization problem we solve with our proposal fully considers the communication time, the memory usage of all robots, and the average distance to the origin, and these aspects were not taken in consideration in \cite{li:2018}.


\subsubsection*{Acknowledgments}
This work was supported by operation Centro-01-0145-FEDER-000019 - C4 - Centro de Compet\^{e}ncias em Cloud Computing, cofinanced by the European Regional Development Fund (ERDF) through the Programa Operacional Regional do Centro (Centro 2020), in the scope of the Sistema de Apoio \`{a} Investiga\c{c}\~{a}o Cientif\'{i}ca e Tecnol\'{o}gica - Programas Integrados de IC\&DT. This work was supported by NOVA LINCS (UIDB/04516/2020) with the financial support of FCT-Funda\c{c}\~{a}o para a Ci\^{e}ncia e a Tecnologia, through national funds.

\subsubsection*{Conflict of Interest}
The authors have no known conflicts of interest beyond the ones related to organisations they belong to, which are Universidade da Beira Interior and C4 - Cloud Computing Competence Centre (C4-UBI).

\bibliographystyle{apalike}
\bibliography{sample}
\appendix
\newpage
\pagenumbering{arabic}
\section{Appendix: Supplemental Material}\label{sup}

Here we give some examples that explain how our proposed methods work by considering the minimization of time and memory independently. Examples of simultaneous minimization of memory and time can be found in the main text.

\subsection{Time}\label{app}
\begin{figure}[!h]\centering
\includegraphics[width=0.3\linewidth]{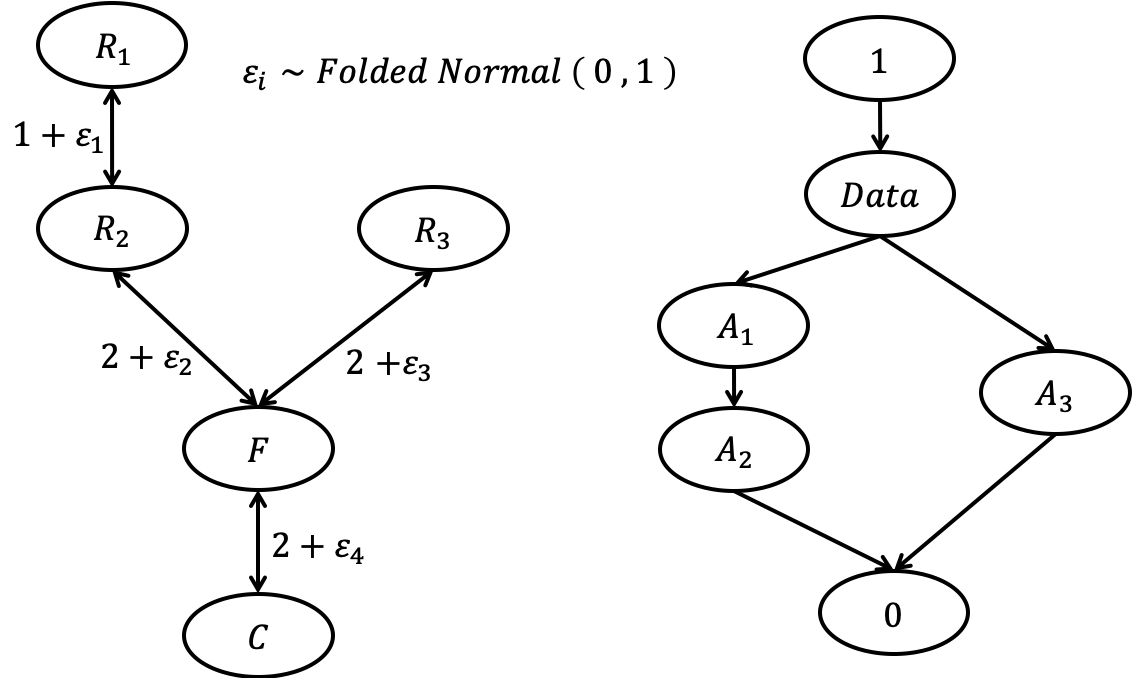}
\caption{Three robots cloud system and algorithms flow. $\mathbf{0}$ and $\mathbf{1}$ are the virtual algorithms. The numbers on the edges are the average data transmission times in seconds, and $\varepsilon_i$'s are random delays in seconds from the folded normal distribution with mean $0$ and variance $1$.}
\label{fig4}
\end{figure}

\begin{figure}[!h]\centering
\includegraphics[width=0.5\linewidth]{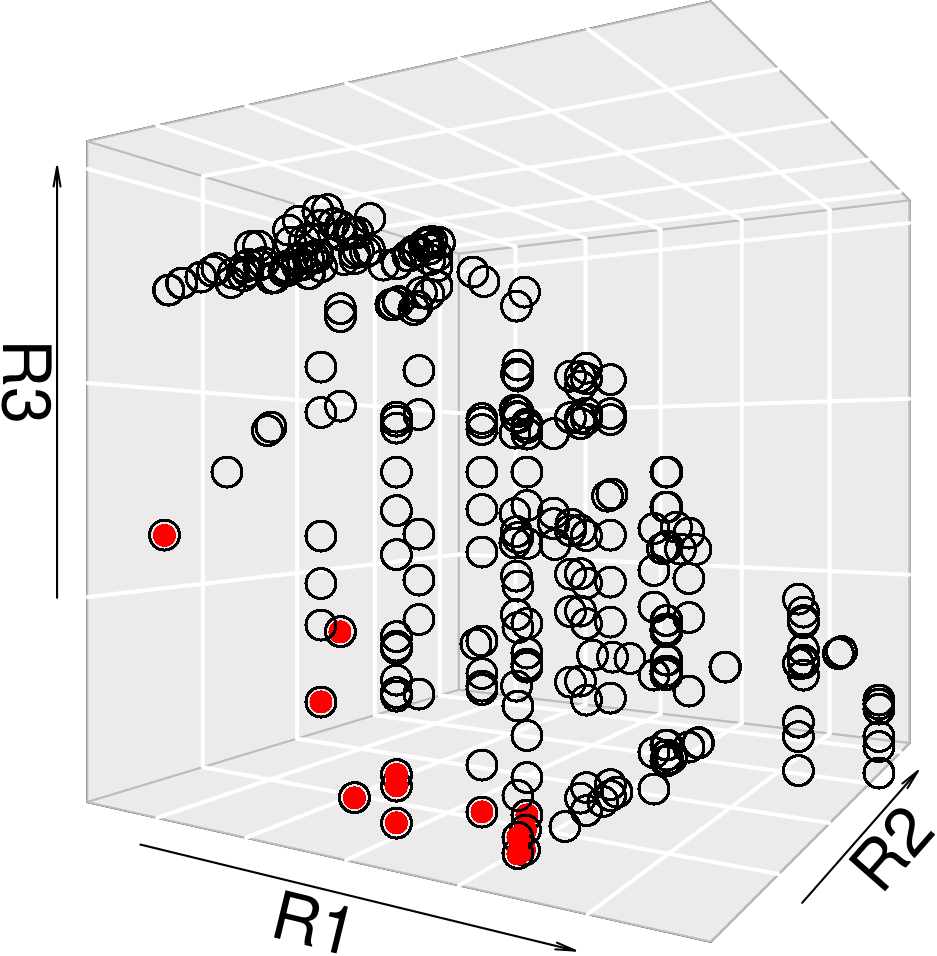}
\caption{Optimized solution for the test with the system and algorithms defined in Figure~\ref{fig4}. For each allocation of algorithms, the times at which the outputs of all algorithms are available for robots 1, 2, and 3 are determined and represented as a point in a $3$-dimensional space. The axis $R_i$ represents the time value for robot $i$, for $i=1,2,3$ in seonds.} 
\label{fig5}
\end{figure}

We performed a test for three robots, see Figure~\ref{fig4}, where the data transmission is data at once, and we added some random delay to the data transmission times. We use $3$ robots because we can have a visualization for it in a three-dimensional space. The solutions of the test are the red dots in Figure~\ref{fig5} and the respective average response times and allocation of the algorithms are shown in Table~\ref{tab2}. The average execution time of each algorithm at each node of the cloud system is shown in Table~\ref{tab1}. We assume that the robots have identical processing power.

\begin{table}[!h]
\caption{The average execution time (in seconds) of algorithm nodes in cloud system's nodes. The average execution time of Data is assumed to be $0$.}
\begin{center}
\begin{tabular}{l|ccc}
&$A_1$&$A_2$&$A_3$\\
\hline
Edge (Robot 1)&2&6&4\\
Edge (Robot 2)&2&6&4\\
Edge (Robot 3)&2&6&4\\
Fog $F$&1&3&2\\
Cloud $C$&0.5&1.5&1\\
\end{tabular}
\end{center}
\label{tab1}
\end{table}

\begin{table*}[!h]
\caption{Optimal average response time.}
\begin{center}
\begin{tabular}{cccccc}
&Edge&&Fog&Cloud&Average Time (s)\\\cline{1-3}
Robot 1&Robot 2&Robot 3&&&\\
\hline
$A_2$&$A_3$&-&Data and $A_1$&-&\textbf{37.87}\\
-&Data, $A_1$ and $A_2$&-&$A_3$&-&37.89\\
-&Data, $A_2$ and $A_3$&-&$A_1$&-&38.12\\
-&$A_1$&-&Data, $A_2$ and $A_3$&-&38.51\\
$A_2$&Data and $A_1$&-&$A_3$&-&38.60\\
-&Data and $A_1$&-&$A_2$ and $A_3$&-&38.69\\
$A_1$&Data&-&$A_2$ and $A_3$&-&38.71\\
-&Data, $A_1$ and $A_3$&-&$A_2$&-&39.34\\
$A_1$ and $A_2$&-&-&Data and $A_3$&-&39.77\\
-&Data&-&$A_1$ and $A_3$&$A_2$&39.94\\
\end{tabular}
\end{center}
\label{tab2}
\end{table*}

In Table~\ref{tab2}, we consider the algorithms' allocation that has an average response time of less than 40 seconds. 

\begin{figure}[tb]\centering
\includegraphics[width=0.5\linewidth]{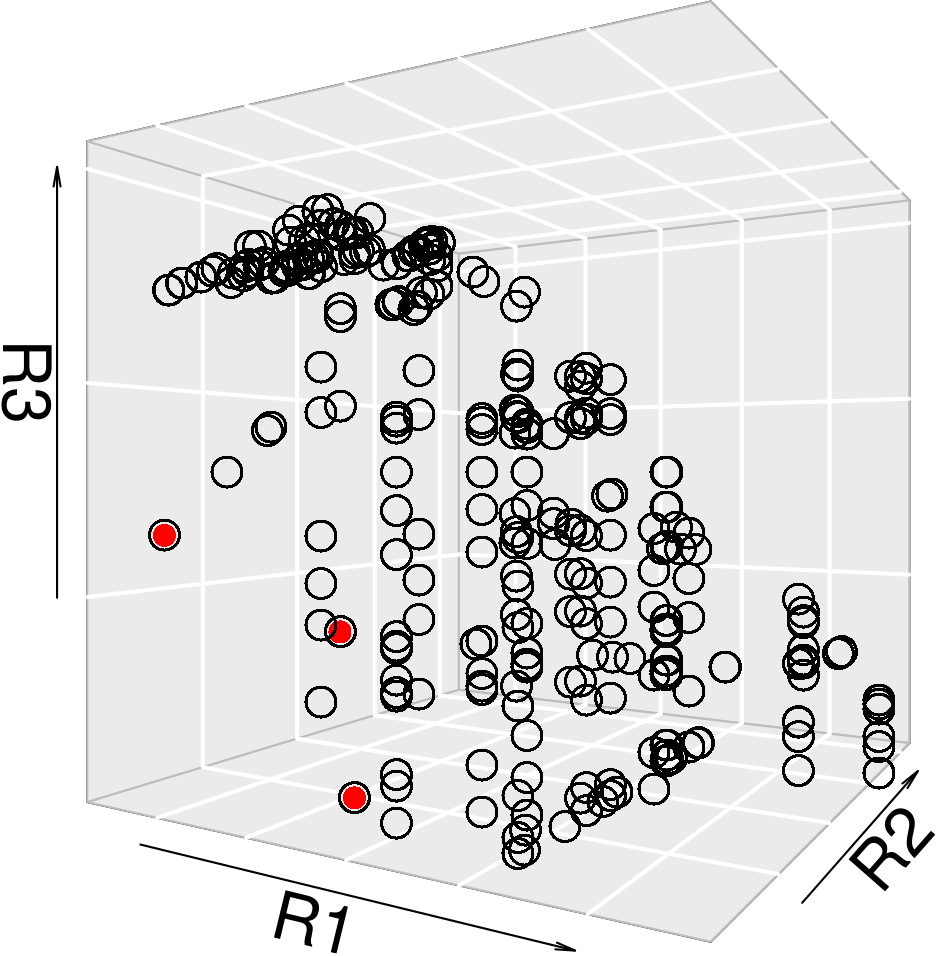}
\caption{Optimized solution for the test with the system and algorithms defined in Figure~\ref{fig4} assuming that $Data$ can only be stored in the cloud or in the fog. Points and axis are the same as in Figure~\ref{fig5}.} 
\label{fig6}
\end{figure}

Now we assume the prior information that $Data$ can only be stored in the cloud or in the fog. The solutions under this assumption are shown as red dots in Figure~\ref{fig6}, and the respective average response times and allocation of algorithms are listed in Table~\ref{tab3}. Again, in Table~\ref{tab3}, we consider the algorithms' allocation such that their average response time is less than 40 seconds. 
\begin{table*}[tb]
\caption{Optimal average response time assuming data can only be stored in the cloud or fog, the red dots.}
\begin{center}
\begin{tabular}{cccccc}
&Edge&&Fog&Cloud&Average Time (s)\\\cline{1-3}
Robot 1&Robot 2&Robot 3&&&\\
\hline
$A_2$&$A_3$&-&Data and $A_1$&-&\textbf{37.87}\\
-&$A_1$&-&Data, $A_2$ and $A_3$&-&38.51\\
$A_1$ and $A_2$&-&-&Data and $A_3$&-&39.77\\
\end{tabular}
\end{center}
\label{tab3}
\end{table*}

\subsection{Memory}

In the first example, for simplicity, we assume simple addition rather than the algebra of memory to reduce additional computational complexity. We also assume that the space complexity of all algorithms is identical.

We performed a test for $5$ robots with $2$ in the class $TR_0=\{\text{Robot 1},\text{Robot 2}\}$. A total of $13$ algorithms are to be executed on the edge nodes, with total memory usage (in $Mbytes$):
$$\{a_1=4, a_2=5, a_3=6, a_4=7, a_5=8, a_6=10, a_7=12, a_8=13, a_9=14, a_{10}=18, a_{11}=24, a_{12}=30, a_{13}=32 \}$$ 
and 
$$\{a_2=5,a_3=6,a_9=14,a_{10}=18,a_{11}=24\}$$ 
are the algorithms with non-zero input memory from the fog or the cloud. By applying Proposition~\ref{prop:prop1}, we have 
\begin{flalign*}
&\text{Robot 1}: a_2=5,a_3=6,a_{11}=24,~~\text{Total}=35 (Mbytes),&&\\
&\text{Robot 2}: a_1=4,a_9=14,a_{10}=18,\text{Total}=36 (Mbytes),&&\\
&\text{Robot 3}: a_4=7,a_{13}=32,\qquad\quad~~\textbf{\text{Total}=39 (Mbytes)},&&\\
&\text{Robot 4}: a_5=8,a_{12}=30,\qquad\quad~~\text{Total}=38 (Mbytes),&&\\
&\text{Robot 5}: a_6=10,a_7=12,a_8=13,\text{Total}=35 (Mbytes).&&
\end{flalign*}

So, the minimum memory requirement for edge nodes is $39$, in addition to the total memory used for all the outputs. Note that the first two and the last three can be swapped, and $a_1$ and $a_4$ can also swap places, and the same optimal result is obtained.

\begin{figure}[tb]\centering
\includegraphics[width=0.5\linewidth]{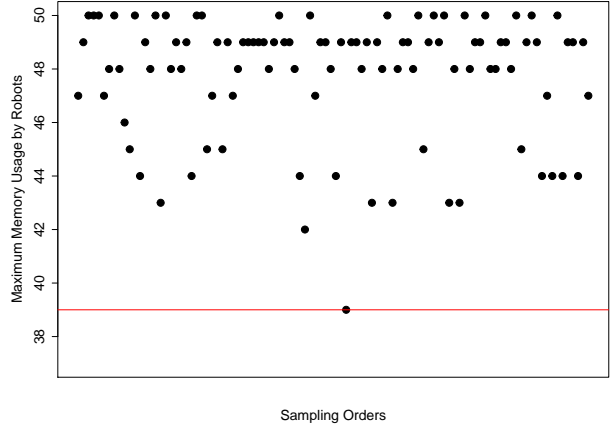}
\caption{Random allocation of algorithms to the edge nodes. The horizontal line is the solution for the minimum memory requirement of the edge nodes (in $Mbytes$). Each dot represents the minimum memory requirement of the edge nodes under random allocation (first $100$ under random allocation that minimum memory requirement of the edge nodes is less than $50$ ($Mbytes$)).}
\label{fig4p}
\end{figure}

Now, we perform a random allocation of the algorithms to the edge nodes. In Figure~\ref{fig4p}, we only plot the first $100$ random allocation of algorithms to edge nodes where the minimum required memory for edge nodes is less than $50$ ($Mbytes$). The point on the horizontal line is one of the equivalent representatives of those shown above.

The following example is intended to explain the minimal memory allocation algorithms in terms of the graph of algorithms and the algebra of memory as in Remark~\ref{rem:rem1}.
\begin{example}
Consider $4$ robots with $2$ in class $TR_0=\{\text{Robot 1},\text{Robot 2}\}$. Assume that a total of $10$ algorithms are to be executed on the edge nodes, and that algorithms $A_1$ and $A_5$ are the algorithms with nonzero input memory from the fog or cloud. Figure~\ref{fig5p} shows the graph of algorithms, and Table~\ref{tb1} shows the memory usage by all algorithms and all possible solutions for the algorithms allocation with minimum necessary memory usage equal to $22$ ($Mbytes$). The algebra of the memory is shown in Figure~\ref{fig6p}. We assume that the memory usage by $m_{ou}$ of all the algorithms are considered in the edge nodes, $m_{in}$ of all the algorithms except the two $A_1$ and $A_5$, which are also considered as the outputs of their ancestors in the graph of algorithms.

\begin{figure}[tb]\centering
\includegraphics[width=0.3\linewidth]{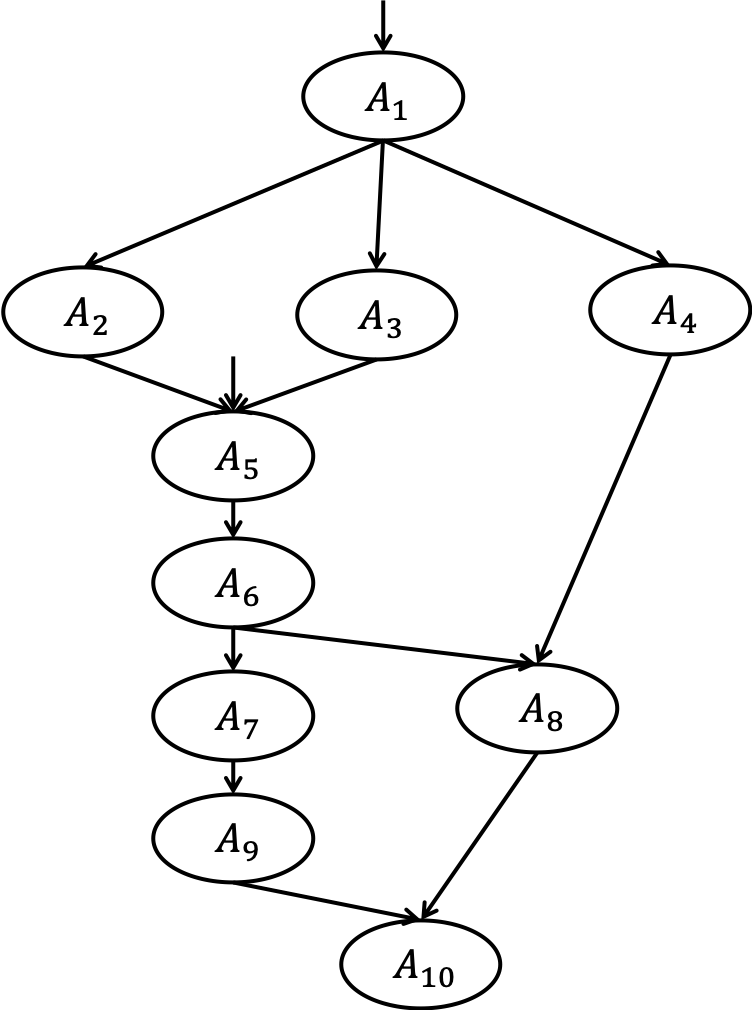}
\caption{Graph of algorithms. The additional incoming arrows to the algorithms $A_1$ and $A_5$ are used to inform that $m_{in}(A_1)|_{F\cup C}\neq0$ and $m_{in}(A_5)|_{F\cup C}\neq0$.}
\label{fig5p}
\end{figure}

\begin{table}[tb]
\caption{Processing and input memory usage by algorithms (in $Mbytes$), and all possible solutions for the algorithms allocation with minimum necessary memory usage are shown.}
\begin{center}
\begin{tabular}{cccl}
Algorithms&$m_{in}|_{F\cup C}$&$m_{pr}$&Algorithm Allocation\\
\hline
$A_1$&10&5&Robot 1\\
$A_2$&-&6&Robots 1, 2, 3, and 4\\
$A_3$&-&12&Robots 1, 3, and 4\\
$A_4$&-&7&Robots 1 and 3\\
$A_5$&11&9&Robot 2\\
$A_6$&-&18&Robots 3 and 4\\
$A_7$&-&20&Robot 4\\
$A_8$&-&22&Robot 3\\
$A_9$&-&14&Robots 1, 3, and 4\\
$A_{10}$&-&8&Robots 1, 2, 3, and 4
\end{tabular}
\end{center}
\label{tb1}
\end{table}

\begin{figure}[tb]\centering
\includegraphics[width=0.7\linewidth]{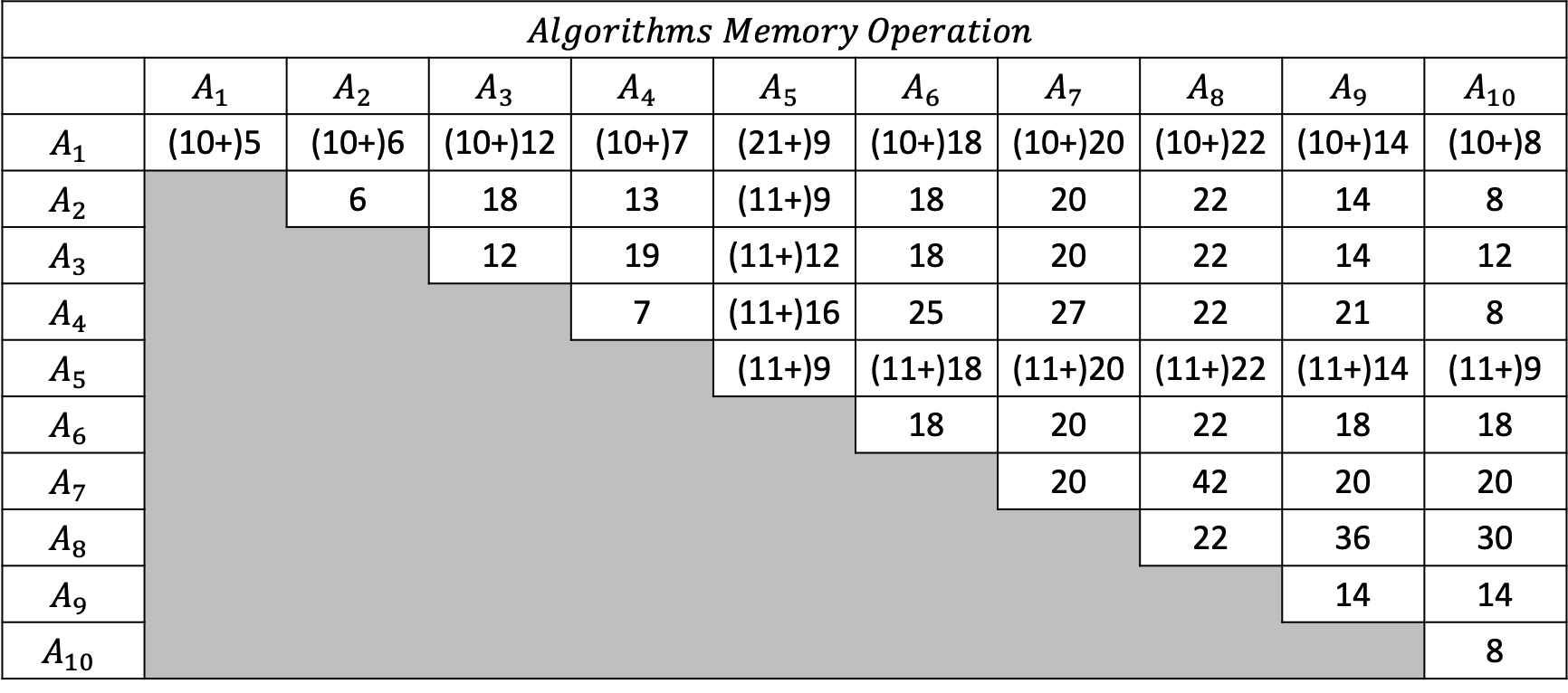}
\caption{Algebra of memory usage by algorithms. For example, $A_4$ and $A_5$ are parallel algorithms, so their processing will be the sum of processing memory of both and equal to $16$ ($Mbytes$), since their processing memories are $7$ ($Mbytes$) and $9$ ($Mbytes$) respectively, and it requires an additional $11$ ($Mbytes$) for the input memory of $A_5$. Moreover, $A_1$ and $A_5$ are serial, so their processing will be the maximum of the processing memory of both and equal to $9$ ($Mbytes$) since their processing memories are $5$ ($Mbytes$) and $9$ ($Mbytes$) respectively, and it needs an additional $21$ ($Mbytes$) for the input memories of $A_1$ and $A_5$ (assuming that their input memories have no intersection).}
\label{fig6p}
\end{figure}
Note that, in Figure~\ref{fig6p}, we only considered the upper triangular operations because the operation is commutative. 

Similar to the previous test, we performed a random allocation of the algorithms on the edge nodes. In Figure~\ref{fig7}, we only plot the first $100$ random allocation of algorithms to edge nodes such that the minimum required memory for the edge nodes is less than $30$ ($Mbytes$). Here, we evaluate the algebra of memory instead of simple addition with respect to the graph of algorithms in Figure~\ref{fig5p} and Table~\ref{tb1}.
\begin{figure}[tb]\centering
\includegraphics[width=0.5\linewidth]{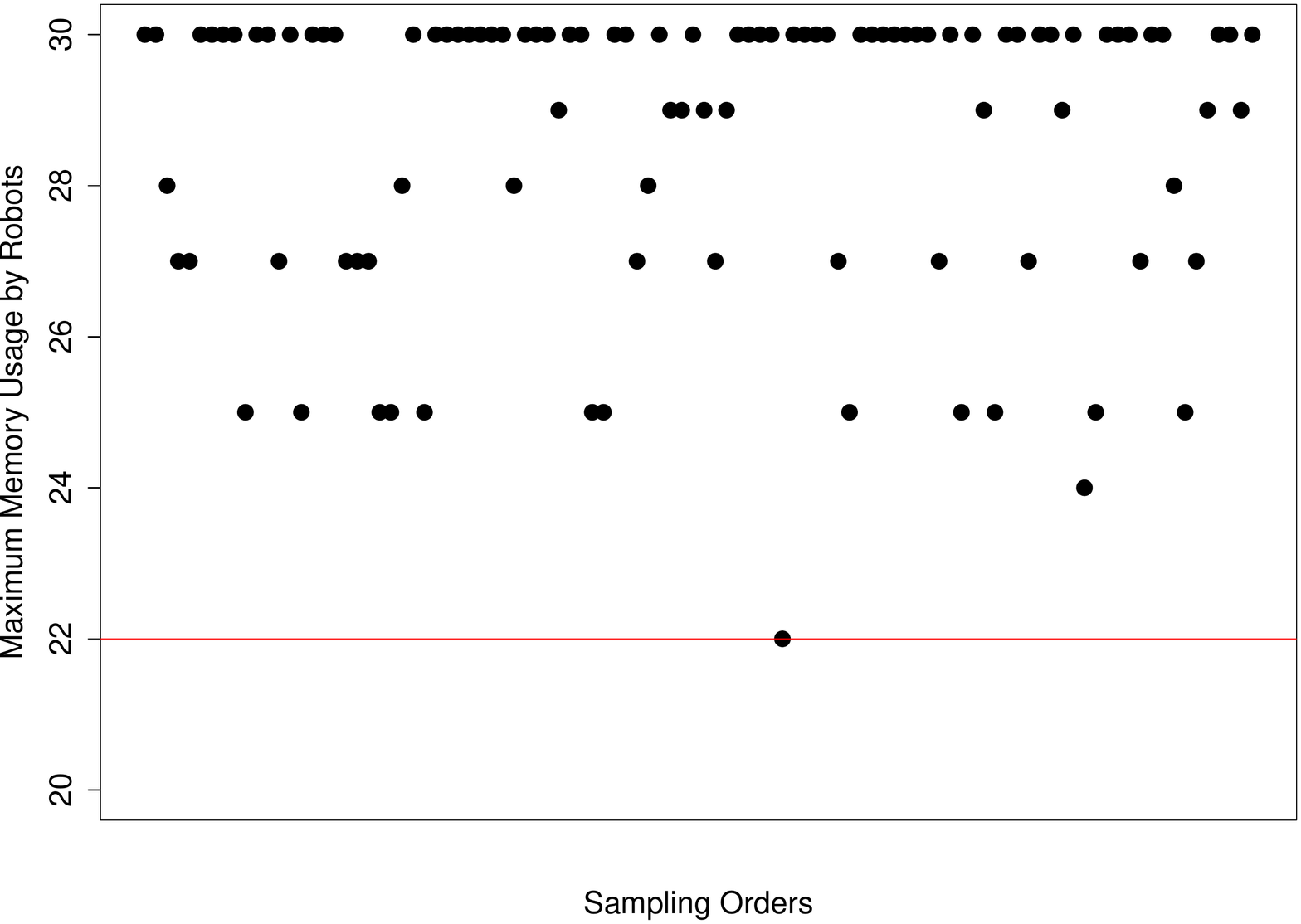}
\caption{Random allocation of the algorithms to edge nodes. The horizontal line is the solution for the minimum memory necessary for edge nodes (in $Mbytes$) and each dot represents the minimum memory requirement for edge nodes with random allocation (first $100$ with random allocation that the minimum memory requirement for edge nodes is less than $30$ ($Mbytes$), using the memory algebra).}
\label{fig7}
\end{figure}
\end{example}

These examples help to understand how our method can be used independently for time and memory components and how to apply it.
\subsection{Exact Values of Experimental Results}
Table~\ref{tab4} shows the exact values for the average distance to the origin and their standard deviations obtained using our method and the method proposed in \cite{li:2018} for randomly generated architectures. The plot of the results can be seen in Figure~\ref{figres}.
\begin{table}[tb]
\caption{Comparison of the distance to the origin for optimal algorithm allocation using our method and the method proposed in \cite{li:2018}, for randomly generated architectures with $n=1,\ldots,10$ robots.}\label{tab4}
\begin{center}
\begin{tabular}{ccc}
Number&\multicolumn{2}{c}{Average distance to the origin}\\
of&\multicolumn{2}{c}{(with standard deviation)}\\\cline{2-3}
robots&Ours&\cite{li:2018}\\
\hline
1&$\mathbf{1.083}$ ($1.798e-02$)&$2.904$ ($1.338e-05$)\\
2&$\mathbf{1.219}$ ($0.148$)&$3.004$ ($0.041$)\\
3&$\mathbf{1.264}$ ($0.186$)&$3.058$ ($0.053$)\\
4&$\mathbf{1.258}$ ($0.143$)&$3.074$ ($0.077$)\\
5&$\mathbf{1.429}$ ($0.064$)&$3.267$ ($0.186$)\\
6&$\mathbf{1.394}$ ($0.049$)&$3.202$ ($0.259$)\\
7&$\mathbf{1.363}$ ($0.095$)&$3.124$ ($0.166$)\\
8&$\mathbf{1.394}$ ($0.127$)&$3.256$ ($0.388$)\\
9&$\mathbf{1.389}$ ($0.104$)&$3.118$ ($0.123$)\\
10&$\mathbf{1.350}$ ($0.103$)&$3.138$ ($0.135$)
\end{tabular}
\end{center}
\end{table}

\subsection{Exact Values of Experimental Results}
Table~\ref{tabtime} shows the exact values for the average execution time of the proposed algorithm for randomly generated architectures. The plot of the results can be seen in Figure~\ref{figtime}.
\begin{table}[tb]
\caption{Average execution time (in seconds) of the proposed method to find the optimal algorithm allocation, for randomly generated architectures with $n=1,\ldots,10$ robots in seconds.}\label{tabtime}
\begin{center}
\begin{tabular}{cc}
Number of robots&Average execution time\\
\hline
1&$0.368$\\
2&$2.93$\\
3&$11.93$\\
4&$40.87$\\
5&$121.19$\\
6&$231.52$\\
7&$504.35$\\
8&$823.80$\\
9&$1204.32$\\
10&$1980.73$
\end{tabular}
\end{center}
\end{table}

\end{document}